\documentclass[sigconf, nonacm, screen]{acmart}

\usepackage{multirow}
\usepackage{enumitem}
\usepackage{subcaption}
\usepackage{color-edits}
\usepackage[dvipsnames]{xcolor}

\addauthor[Sergio]{sj}{teal}

\begin{document}

\title[A Human-Centered Audit of Shapley Value Benchmarks in High-Stakes Workflows]{Rethinking XAI Evaluation: A Human-Centered Audit of Shapley Benchmarks in High-Stakes Settings}


\author{In\^es Oliveira e Silva}
\orcid{0009-0002-2395-298X}
\affiliation{%
  \institution{University of Porto, Feedzai}
  \city{Porto}
  \country{Portugal}
}
\email{ines.silva@feedzai.com}

\author{Sérgio Jesus}
\affiliation{%
  \institution{Feedzai}
  \city{Porto}
  \country{Portugal}
}
\email{sergio.jesus@feedzai.com}

\author{Iker Perez}
\orcid{0000-0001-9400-4229}
\affiliation{%
  \institution{Feedzai}
  \city{London}
  \country{UK}
}
\email{iker.perez@feedzai.com}

\author{Rita P. Ribeiro}
\affiliation{%
  \institution{University of Porto}
  \city{Porto}
  \country{Portugal}
}
\email{rpribeiro@fc.up.pt}

\author{Carlos Soares}
\affiliation{%
  \institution{University of Porto}
  \city{Porto}
  \country{Portugal}
}
\email{csoares@fe.up.pt}

\author{Hugo Ferreira}
\affiliation{%
  \institution{Feedzai}
  \city{Lisbon}
  \country{Portugal}
}
\email{hugo.ferreira@feedzai.com}

\author{Pedro Bizarro}
\affiliation{%
  \institution{Feedzai}
  \city{Lisbon}
  \country{Portugal}
}
\email{pedro.bizarro@feedzai.com}

\renewcommand{\shortauthors}{Oliveira e Silva et al.}

\begin{abstract}
Shapley values are a cornerstone of explainable AI, yet their proliferation into competing formulations has created a fragmented landscape with little consensus on practical deployment. While theoretical differences are well-documented, evaluation remains reliant on quantitative proxies whose alignment with human utility is unverified. In this work, we use a unified amortized framework to isolate semantic differences between eight Shapley variants under the low-latency constraints of operational risk workflows. We conduct a large-scale empirical evaluation across four risk datasets and a realistic fraud-detection environment involving professional analysts and 3,735 case reviews. Our results reveal a fundamental misalignment: standard quantitative metrics, such as sparsity and faithfulness, are decoupled from human-perceived clarity and decision utility. Furthermore, while no formulation improved objective analyst performance, explanations consistently increased decision confidence, signaling a critical risk of automation bias in high-stakes settings. These findings suggest that current evaluation proxies are insufficient for predicting downstream human impact, and we provide evidence-based guidance for selecting formulations and metrics in operational decision systems.
\end{abstract}

\keywords{Explainable AI (XAI), Shapley values, Feature attribution, Evaluation methodology, Human-in-the-loop, Automation bias}
\maketitle

\section{Introduction}
\label{sec:introduction}

\begin{figure*}[t]
    \centering
    \begin{subfigure}[t]{0.31\textwidth}
        \centering
        \includegraphics[width=\textwidth]{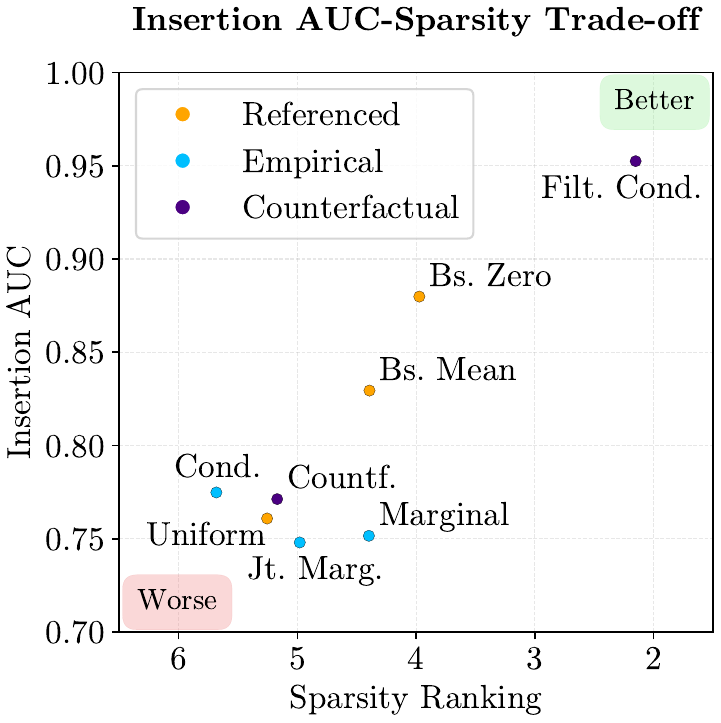}
        \label{fig:teaser-quant}
    \end{subfigure}
    \hfill
    \begin{subfigure}[t]{0.31\textwidth}
        \centering
        \includegraphics[width=\textwidth]{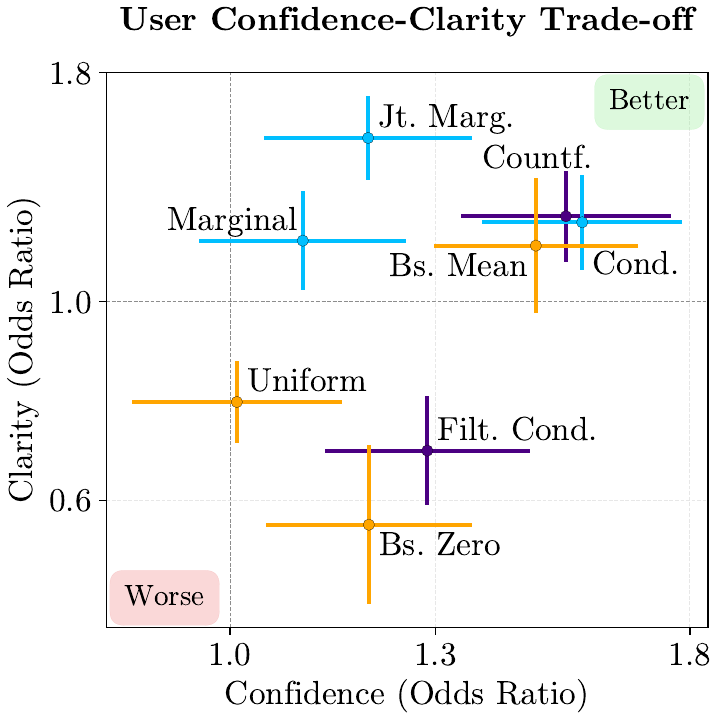}
        \label{fig:teaser-human}
    \end{subfigure}
    \begin{subfigure}[t]{0.34\textwidth}
        \centering
        \includegraphics[width=\textwidth]{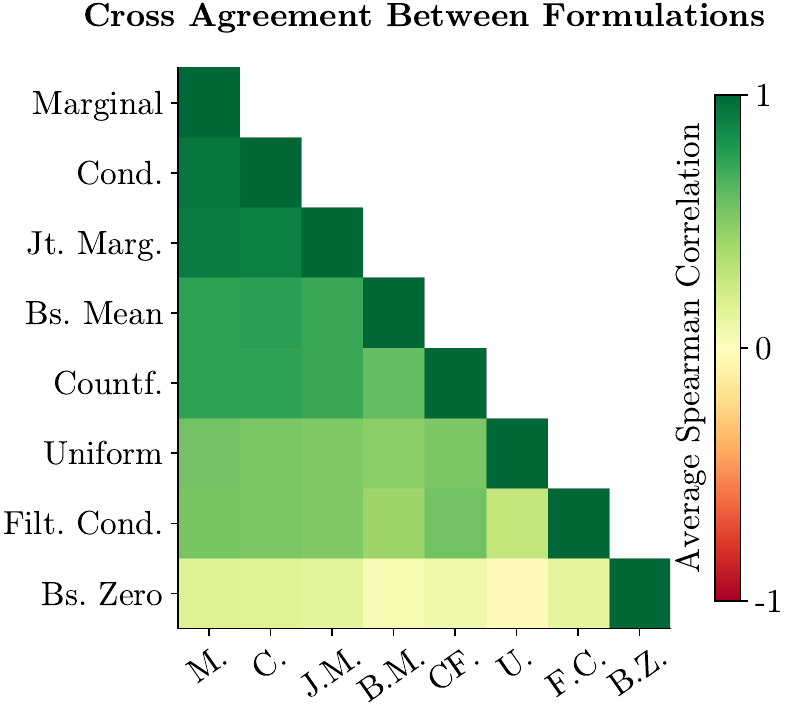}
        \label{fig:teaser-stability}
    \end{subfigure}
    \hfill
    \caption{
    Overview of the empirical XAI audit in this paper.
    \textbf{Left:} Quantitative benchmarks reveal systematic functional variation (e.g., sparsity, faithfulness) across Shapley formulations. 
    \textbf{Center:} Controlled analyst studies show differences in perceived clarity and decision confidence.
    \textbf{Right:} Pairwise agreement between Shapley formulations reveals structured alignment.
    }
    \label{fig:teaser}
\end{figure*}

\textit{Machine learning} (ML) systems are increasingly deployed in high-stakes domains such as fraud detection, credit assessment, and healthcare, where predictions have direct human and financial consequences \cite{khan2025model, mienye2024survey}. In these settings, model outputs rarely constitute final decisions. Instead, predictions are reviewed by human decision-makers operating under time, attention, and regulatory constraints. As a result, explanations are viewed as indispensable for accountability and oversight, and have become a core requirement in operational ML deployments \cite{bucker2022transparency, bhatt2020explainable, rudin2019stop}. Yet, despite widespread adoption, the practical value of explanations in human-in-the-loop workflows remains poorly understood, often assumed rather than empirically established \cite{bhatt2020explainable, jesus2021how, amarasinghe2024importance}.

Among explanation methods, local approaches grounded in cooperative game theory, most notably \emph{Shapley values}, have emerged as the \textit{de facto} standard for feature attribution~\cite{ali2023explainable}, by providing an axiomatic decomposition of model predictions into feature-level contributions \cite{lundberg2017unified}. However, the framework has fragmented into competing formulations based on divergent assumptions about the semantics of feature absence, realized in popular implementations such as KernelSHAP, TreeSHAP, and related tools \cite{zern2023interventional, albini2022counterfactual, chen2023algorithms, olsen2024improving, covert2021improving, lundberg2018consistent, yang2021fast}. This raises a critical evaluation question for practitioners: \emph{does the choice of formulation matter to the end-user, and do standard evaluation procedures anticipate the impact?}

Modern XAI evaluation relies on theoretical analysis and quantitative proxies \cite{merrick2020explanation, covert2021explaining, chen2023algorithms}, as well as mathematical distinctions between ``faithfulness'' to the model or ``truthfulness'' to the data \cite{chen2020true}. Yet, systematic evidence regarding how explanation methods perform against human-centered benchmarks remains scarce. Existing evaluations often focus on isolated \textit{functional} (model-based) properties and rarely stress-test these metrics against human behavior under realistic operational constraints \cite{nauta2023anecdotal, canha2025functionally}. Furthermore, comparisons are often confounded by implementation choices, which mask the true semantic differences of the definitions themselves.

In this work, we treat XAI evaluation as a scientific object of study. We use Shapley values as a representative class of feature attribution, and audit the alignment between quantitative evaluation benchmarks and human utility through a comprehensive study of 3,735 instances of human-AI interaction. We use a unified amortized framework to eliminate implementation confounders, and instantiate eight Shapley definitions under identical computational conditions. Our scope reflects high-throughput, high-stakes tabular data environments in low-latency financial, healthcare, e-commerce, or logistics deployments \cite{sahakyan2021explainable, grinsztajn2022tree, pmlr-v235-van-breugel24a}. We conduct this audit across 5 datasets in a real production-grade fraud-detection environment involving 37 participants, including professional analysts.



Our analysis reveals a fundamental \textbf{decoupling between metrics and utility}: standard quantitative proxies are poor predictors of human-perceived clarity or decision confidence. Crucially, while explanations fail to improve objective analyst performance, they consistently boost confidence. This exposes a major deployment risk: current evaluation practices may favor explanations that encourage \textbf{automation bias} without improving decision quality. In Figure~\ref{fig:teaser} we provide an overview of the landscape explored in this paper. Our contributions are:
\begin{itemize}
    \item In Section~\ref{sec:foundations}, we describe a unified framework to isolate semantic explanation effects of Shapley formulations, eliminating algorithmic confounders for fair comparison.
    \item In Sections~\ref{sec:evaluation}-\ref{sec:results}, we demonstrate through expert-driven reviews that common XAI evaluation metrics fail to align with human decision-making outcomes.
    \item In Section~\ref{sec:discussion}, we document a critical failure mode: the promotion of decision confidence without accuracy gains. 
\end{itemize}
We further release 3,735 granular human-AI interaction measurements to support the development of behaviorally grounded XAI benchmarks.

\renewcommand{\arraystretch}{1.25}
\setlength{\tabcolsep}{3.5pt}
\begin{table*}[t!]
    \centering
    \caption{Taxonomy of Shapley value formulations categorized by the representation of feature absence.}
    \begin{tabular}{p{2.1cm} p{3.1cm} p{3.5cm} p{8.1cm}}
    \toprule
    \textbf{Category} & \textbf{Definition} & \textbf{Sampling Distribution} & \textbf{Semantics / Assumptions} \\
    \midrule

    Referenced
    & Fixed Baseline
    & $\delta(\boldsymbol{X}_{\mathcal{F} \setminus \mathcal{S}} = \boldsymbol{x}'_{\mathcal{F} \setminus \mathcal{S}})$
    & Absent features fixed to a reference; efficient but highly sensitive to baseline selection. \\

    & Uniform
    & $u(\boldsymbol{X}_{\mathcal{F} \setminus \mathcal{S}})$
    & Samples uniformly from a hyperbox; ignores data distribution and feature correlations. \\

    Empirical
    & Marginal
    & $p(\boldsymbol{X}_{\mathcal{F}\setminus\mathcal{S}})$
    & Samples absent features from joint empirical distribution; preserves marginals but breaks feature dependencies with $\boldsymbol{x}_{\mathcal{S}}$. \\

    & Joint-Marginal
    & $\prod_{j\in\mathcal{F}\setminus\mathcal{S}} p(X_j)$
    & Samples from univariate marginals independently; removes all feature dependencies. \\

    & Conditional
    & $p(\boldsymbol{X}_{\mathcal{F}\setminus\mathcal{S}} \mid \boldsymbol{X}_{\mathcal{S}}=\boldsymbol{x}_{\mathcal{S}})$
    & Preserves empirical dependencies via conditioning; difficult to estimate in high-dimensional settings. \\

    Counterfactual
    & Search Counterfactual
    & $p(\boldsymbol{X}^{(c)}_{\mathcal{F}\setminus\mathcal{S}} \mid f(\boldsymbol{X}_{\mathcal{F}}^{(c)})\!\approx\!y^*)$
    & Uses counterfactual perturbations as baselines; high variance and optimization-intensive. \\

    & Filtered Conditional
    & $p(\boldsymbol{X}_{\mathcal{F}\setminus\mathcal{S}} \mid f(\boldsymbol{X}_{\mathcal{F}})\!\in\!\mathcal{Y})$
    & Conditions on background samples with specific model outputs; blends attribution with contrastive logic. \\

    \bottomrule
    \end{tabular}
    \label{tab:shapdefinitions}
\end{table*}

\section{Shapley Value Foundations}
\label{sec:foundations}

Shapley values~\cite{shapley1953value} provide an additive decomposition of a model’s output
$f:\mathcal{X}\!\mapsto\!\mathbb{R}$ (e.g., risk scores or regression estimates) into feature-level contributions. Their axiomatic foundation and status as \textit{de facto} standard for tabular applications is ideal for auditing alignment between XAI evaluation proxies and human utility. 

Let $\mathcal{F}=\{1,\ldots,d\}$ denote the full feature set and $\mathcal{S}\!\subseteq\!\mathcal{F}$ a coalition of ``present'' features. For an input instance $\boldsymbol{x}\in\mathcal{X}$, a \emph{value function} $v_{\boldsymbol{x}}(\mathcal{S})$ specifies the expected model output when only features in $\mathcal{S}$ are observed:
\begin{equation}
v_{\boldsymbol{x}}(\mathcal{S}) = 
\mathbb{E}_{\boldsymbol{X}_{\mathcal{F}\setminus \mathcal{S}} \sim p(\cdot)}
\big[ f(\boldsymbol{x}_{\mathcal{S}}, \boldsymbol{X}_{\mathcal{F}\setminus \mathcal{S}}) \big],
\label{eq:valuefunction_general}
\end{equation}
where the \textit{background} distribution $p(\cdot)$ defines the semantics of feature absence. The Shapley contribution of feature $i$ to a model output is defined as
\begin{equation}
\phi_i = \sum_{\mathcal{S}\subseteq \mathcal{F}\setminus\{i\}}
\frac{|\mathcal{S}|!(d-|\mathcal{S}|-1)!}{d!}
\big[v_{\boldsymbol{x}}(\mathcal{S}\cup\{i\})-v_{\boldsymbol{x}}(\mathcal{S})\big],
\label{eq:shapley}
\end{equation}
which satisfies efficiency, symmetry, and dummy axioms~\cite{lundberg2017unified, sundararajan2020many}.

\paragraph{A Taxonomy of Semantic Variants.}
Equation~\eqref{eq:valuefunction_general} highlights that Shapley values are a family of definitions induced by the choice of background distribution $p(\cdot)$. This choice reflects the tension between being ``true to the model'' versus ``true to the data''~\cite{chen2020true} and can produce divergent attribution patterns~\cite{lundberg2018consistent, albini2022counterfactual, janzing2020feature, datta2016algorithmic} with direct implications for human-in-the-loop workflows~\cite{chen2023algorithms}. While causal~\cite{heskes2020causal} or in-manifold~\cite{taufiq2023manifold} variants also exist, they require generative models or assumptions often incompatible with the low-latency, high-throughput requirements of production tabular systems. We therefore focus our audit on computationally feasible variants prevalent in production settings, summarized in Table~\ref{tab:shapdefinitions}. 
 
\subsection{Amortization as an Experimental Control}
\label{sec:amortisation}

Exact computation of Equation~\eqref{eq:shapley} requires evaluating $2^d$ feature coalitions, which is intractable for all but very small $d$. The \textsc{SHAP} framework~\cite{lundberg2017unified} addresses this by recasting Shapley estimation as fitting an additive surrogate model
\begin{equation*}
g(S) = v_{\boldsymbol{x}}(\emptyset) + \sum_{i\in S}\phi_i,
\end{equation*}
subject to an \textit{efficiency} constraint $f(\boldsymbol{x}) = g(\mathcal{F})$.
In this setting, KernelSHAP~\cite{lundberg2017unified} estimates $\phi_i$ for a given instance $\boldsymbol{x}\in\mathcal{X}$ by solving a weighted least-squares regression over sampled coalitions:
\begin{equation}
\mathop{\arg\min}_{\phi_1,\ldots,\phi_d}
\sum_{S\subseteq\mathcal{F}}\pi(S)\big(v_{\boldsymbol{x}}(S)-g(S)\big)^2,
\label{eq:leastSquares}
\end{equation}
with kernel weights
$$\pi(S)=\frac{d-1}{{d\choose |S|}|S|(d-|S|)}$$
emphasizing small and large coalitions.
In practice, Monte Carlo approximations of Equation~\eqref{eq:leastSquares} introduce variance and bias~\cite{goldwasser2024stabilizing}. Consequently, a fragmented landscape of model-specific heuristic optimizations has emerged \cite{lundberg2018consistent, yang2021fast, bifet2022linear, muschalik2024beyond}, which often confound comparisons between Shapley definitions with algorithmic noise.

To isolate semantic effects from implementation artifacts, we utilize \textbf{surrogates} or \textbf{amortizers}~\cite{jethani2022fastshap, covert2024stochastic} as a unified experimental control. Instead of solving~\eqref{eq:leastSquares} independently for each $\boldsymbol{x}$, a parametric universal approximator $\hat{\phi}_\theta(\boldsymbol{x})$ is trained to minimize the expected attribution loss over the data distribution:
\begin{equation}
\min_{\theta}\ 
\mathbb{E}_{\boldsymbol{X}}\Bigg[
\sum_{S\subseteq\mathcal{F}}\pi(S)
\big(v_{\boldsymbol{X}}(S)-\hat{g}_\theta(S;\boldsymbol{X})\big)^2
\Bigg],
\label{eq:fastshap}
\end{equation}
where
\begin{equation*}
\hat{g}_\theta(S;\boldsymbol{x})
= v_{\boldsymbol{x}}(\emptyset)
+\sum_{i\in S}\hat{\phi}_{i,\theta}(\boldsymbol{x})
\end{equation*}
is an additive model satisfying $\sum_i\hat{\phi}_{i,\theta}(\boldsymbol{x}) = f(\boldsymbol{x})-v_{\boldsymbol{x}}(\emptyset)$. This reduces inference to a single forward pass, enabling the millisecond-level Service Level Agreements (SLAs) required in production risk systems~\cite{meyer2023training, dyer2024interventionally, covert2024stochastic}. Crucially, this separation of computational mechanics from theoretical semantics enables like-for-like comparisons across Shapley formulations under identical conditions.

\section{Auditing XAI Metric Alignment}
\label{sec:evaluation}

Prior research has established broad desiderata for explanation quality~\cite{doshi2017towards, vilone2021notions, saeed2023explainable}, converging on \emph{functionally grounded properties} that characterize faithfulness to the predictive model, robustness to perturbations, selectivity, or truthfulness (see Appendix~\ref{app:functional}). However, these properties are frequently treated as ends in themselves~\cite{canha2025functionally}. In this work, we treat such metrics as \textbf{quantitative proxies subject to audit}, and seek to explore their alignment with actual human utility in high-stakes workflows.

We evaluate Shapley attributions along two complementary axes:
(i) \emph{model-based} quantitative proxies, and
(ii) \emph{utility} in a large-scale human-in-the-loop study. 
To isolate the effect of the underlying \emph{Shapley formulation} as the primary independent variable, we hold the explanation interfaces and interaction mechanisms strictly fixed.
Our study spans classification tasks on benchmark datasets~\cite{ding2021retiring, ahmed2020review, statlog_(german_credit_data)_144, fico2025educational-analytics} and \textbf{real-world financial transactions} reviewed by professional analysts. Non-proprietary code and data are released for reproducibility.\footnote{https://github.com/feedzai/SHAP-Value-Function-Evaluation}

\subsection{Evaluation Metrics and Proxies}
\label{sec:eval_metrics}

We use a compact set of metrics to capture functional properties, cross-formulation agreement, and downstream analyst behavior. 
Metrics are defined per instance $\boldsymbol{x}$, and grounded in established notions such as sensitivity, rank agreement, uncertainty, or sparsity. Results are aggregated across datasets, models, and experimental repetitions.

\subsubsection{Quantitative Evaluation}
\label{sec:quantitativeevaluation}

We evaluate whether attributions exhibit behaviors theoretically associated with faithful and usable explanations, independent of human judgment.

\textit{Deletion AUC} assesses faithfulness by measuring how predictive uncertainty increases as features are removed in order of importance~\cite{perez2022attribution}. Let $\boldsymbol{x}^{(k)}$ denote an input with top-$k$ features removed, and $H(\cdot)$ the predictive entropy. Faithful explanations identify features whose removal quickly degrades model confidence:
\begin{equation}
\label{eq:deletionauc}
    1 - \frac{1}{d}\sum_{k=1}^{d}
    \frac{H(\boldsymbol{x}^{(k)})-H(\boldsymbol{x}^{(0)})}
         {H(\boldsymbol{x}^{(d)})-H(\boldsymbol{x}^{(0)})}.
\end{equation}
An analogous \textit{Insertion AUC} measures confidence gains as features are reintroduced.

\textit{Perturbation Sensitivity} measures local stability relative to model output changes under perturbations $\epsilon>0$:
\begin{equation}
\label{eq:sensitivity}
    \|\hat{\phi}_\theta(\boldsymbol{x}+\epsilon)-\hat{\phi}_\theta(\boldsymbol{x})\|_2 \ \big/ \ \left[|f(\boldsymbol{x}+\epsilon)-f(\boldsymbol{x})|+\delta\right],
\end{equation}
where $\delta>0$ ensures numerical stability. Lower values indicate smoother and robust explanations.

\textit{Counterfactual Contrastivity} evaluates if explanations meaningfully differ across perturbed instances $\boldsymbol{x}'$ that cross the decision boundary:
\begin{equation}
\label{eq:contrastivity}
    \|\hat{\phi}_\theta(\boldsymbol{x})-\hat{\phi}_\theta(\boldsymbol{x}')\|_2 \ \big/ \ \left[|f(\boldsymbol{x})-f(\boldsymbol{x}')|+\delta\right].
\end{equation}
Higher values indicate explanations that better reflect decision-relevant model changes.

\textit{Sparsity} (L$_1$/L$_2$) quantifies the concentration of attribution mass across features, a property assumed to reduce cognitive load:
\begin{equation}
\label{eq:sparsity}
    \|\hat{\phi}_\theta(\boldsymbol{x})\|_1 \ \big/ \ \|\hat{\phi}_\theta(\boldsymbol{x})\|_2.
\end{equation}
A lower ratio indicates selective attributions concentrated on fewer features.

\subsubsection{Amortizer Alignment and Agreement}
\label{sec:amortiseragreementevaluation}

To validate our experimental control, we measure
\begin{itemize}
    \item \textit{Attribution Error}: The mean squared error (MSE) between amortized $\hat{\phi}_{j,\theta}(\boldsymbol{x})$ and a high-sample KernelSHAP ground truth $\phi_j(\boldsymbol{x})$; and
    \item \textit{Recall@$k$}: The overlap across the top-$k$ most influential features between amortized and ground truth attributions.
\end{itemize}
In all cases, KernelSHAP references are tailored to each specific Shapley definition. Finally, we measure \textit{Cross-Agreement} (Spearman correlation) between different Shapley formulations to quantify how the choice of definition alters the explanation rank-order.

\subsubsection{Human-in-the-Loop Utility}
\label{sec:humanevaluation}

To assess operational impact, we measure analyst behavior in tasks where model predictions are paired with different Shapley formulations.
We record \textit{Decision Accuracy} ($\mathbb{I}\{y=\hat{y}\}$) and \textit{Decision Time} ($\log(1+t)$) to measure efficiency. In addition, we collect self-reported \textit{Confidence} and \textit{Clarity} scores.
By analyzing the gap between Confidence and Accuracy, we specifically audit for \textbf{automation bias}, where explanations increase trust without improving decision quality.

\subsection{User Interface and Analyst Workflow}
\label{sec:ui_workflow}

\begin{figure*}[t]
    \centering
    \includegraphics[width=0.98\linewidth]{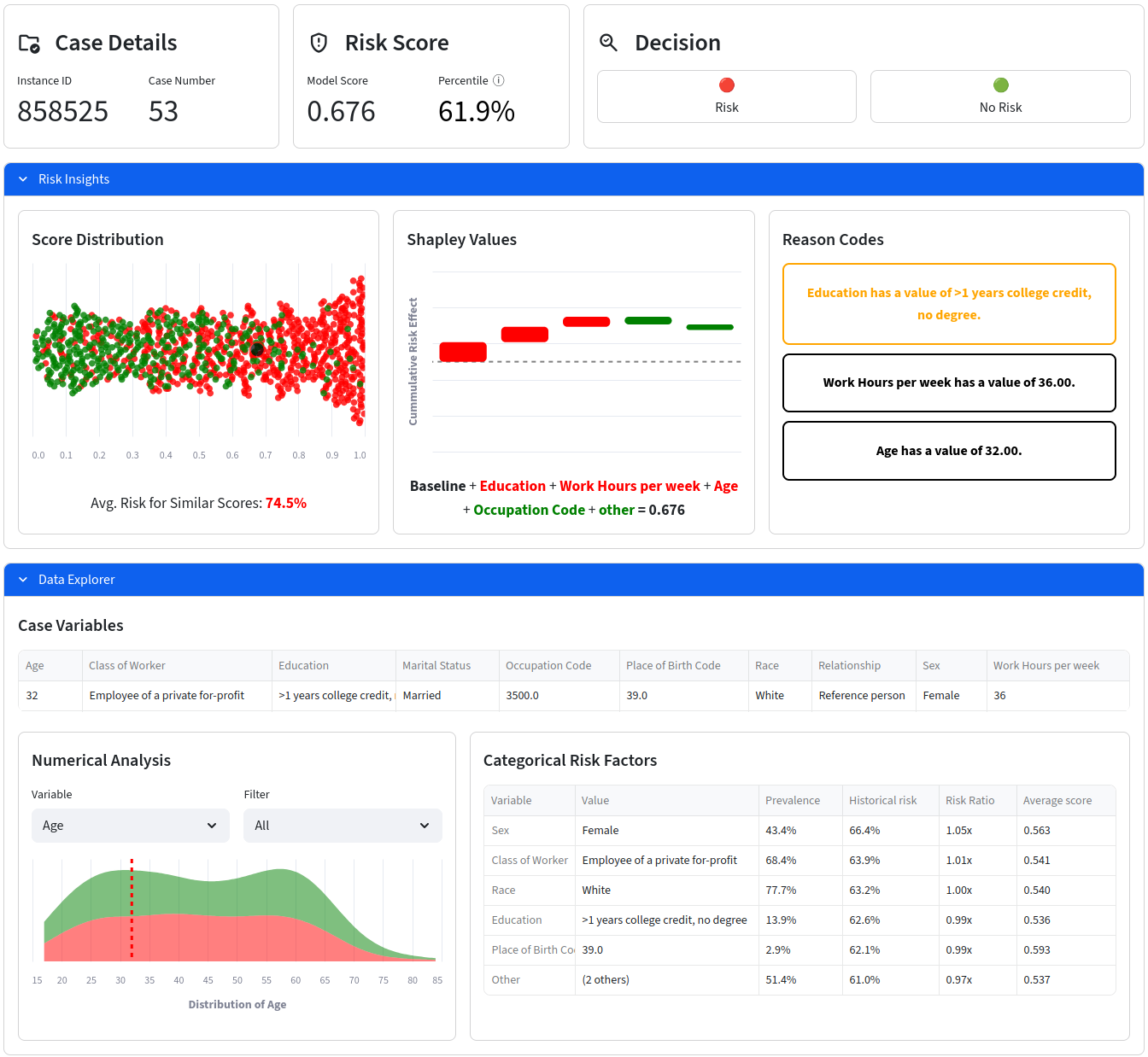}
    \caption{
    The case review interface integrates model scores, Shapley attributions, and contextual data summaries while holding visual encodings fixed across Shapley formulations.}
    \label{fig:main_casereview}
\end{figure*}

Experiments use a standardized interface (Figure~\ref{fig:main_casereview}) grounded in academic conventions~\cite{wysocki2023assessing, lundberg2017unified} and mirroring professional risk tools~\cite{jesus2021how}. It presents a model score/percentile, a Shapley attribution bar chart, and natural-language reason codes derived from the attributions. Visual encodings and interaction affordances are identical across all experimental conditions to ensure observed differences arise solely from the \emph{content} of the Shapley formulation. Further interface details are discussed in Appendix~\ref{app:abtesting}.

Analysts submit a binary decision (\emph{risk}/\emph{no risk}) and self-report confidence (\emph{weak}/\emph{moderate}/\emph{strong}) and clarity (\emph{confusing}/\emph{clear}). Each review produces a structured log of analyst metadata, decision outcome, confidence, timestamps, case features, model predictions, and the Shapley formulation shown.  We utilize two deployments: an open-source frontend for benchmarks and a sandboxed system integrated into real-world fraud review pipelines.

\section{Methodology}
\label{sec:statistics}

To establish a rigorous foundation for our audit, we proceed in two stages: (1) a multi-dimensional evaluation of Shapley formulations across functional and behavioral axes, and (2) a synthesis testing whether quantitative proxies reliably predict human utility.

Our evaluation spans an unusually large-scale experimental matrix: 
(i) \textbf{5 risk datasets}, including Maternal~\cite{ahmed2020review}, Credit~\cite{statlog_(german_credit_data)_144},
HELOC~\cite{fico2025educational-analytics}, Adult~\cite{ding2021retiring}, and a proprietary \textbf{real-world} fraud dataset;
(ii) \textbf{2 industry-standard predictive models} in low-latency systems (Logistic Regression and LightGBM~\cite{ke2017lightgbm}); 
(iii) \textbf{8 distinct Shapley formulations} (Table~\ref{tab:shapdefinitions}); 
and (iv) \textbf{37 analysts}, including professional fraud-review experts. A total of \textbf{3735 case-review measurements} are released publicly\footnote{https://github.com/feedzai/SHAP-Value-Function-Evaluation/tree/main/data} (details in Appendices~\ref{app:datasets}--\ref{app:abtesting}).

\subsection{Aggregation of Quantitative Metrics}
\label{sec:quant_aggregation}

To enable meaningful comparison across heterogeneous tasks, we distinguish between \textit{scale-invariant} and \textit{scale-dependent} metrics. Scale-invariant metrics (e.g., Deletion AUC, Recall@$k$) are averaged directly across dataset--model pairs as they share a consistent interpretation. Conversely, scale-dependent metrics (e.g., sensitivity, sparsity) are converted to relative explainer rankings within each pair before aggregation to ensure cross-task commensurability. All results are reported as point estimates with standard errors obtained via bootstrapping ($50$ subsamples per pair).

\subsection{Human-in-the-Loop Experimental Design}
\label{sec:humanevaluation_design}

Experiments follow a blinded, randomized \textit{within-subjects} design with a no-explanation control. Participants review cases balanced across operational decision thresholds. By holding interface layout and interaction constraints constant, we isolate the Shapley formulation's semantic content as the primary behavioral driver.

\paragraph{Inferential Modeling}
To isolate explanation effects from confounders like case difficulty, analyst expertise, or repeated exposure, we employ \textit{mixed-effects modeling}~\cite{pinheiro2000mixed, fitzmaurice2012applied, nelder1972generalized}. We fit a global model across datasets, predictive models, and analysts, to account for structured heterogeneity and maximize statistical power for generalizable inference. We model effects relative to a no-explanation baseline for \emph{accuracy} (logistic), \emph{confidence} (ordinal), and \emph{time} (log-linear), directly auditing for \textbf{automation bias} where explanations inflate trust without improving decisions. Perceived \emph{clarity} is modeled relative to the population mean, excluding no-explanation cases. Fixed effects include the Shapley formulation, \emph{model type}, \emph{dataset}, \emph{model entropy}, \emph{prediction error}, and \emph{analyst experience}, with further controls for self-reported \emph{subject domain}, \emph{ML}, and \emph{Shapley} familiarity. Results are reported as odds ratios or multiplicative effects with $95\%$ confidence intervals (details in Appendix~\ref{app:abtesting}).

\paragraph{Synthesis}
To bridge quantitative and qualitative metrics, we extend the mixed-effects framework at the \emph{case level}. We use quantitative explanation properties (e.g., sparsity, AUC) as predictors for perceived \textit{clarity} and \textit{confidence}, applying the same rigorous confounder controls. Thus, we statistically determine if offline proxies are reliable predictors of human utility in operational settings.

\section{Results}
\label{sec:results}

Table~\ref{tab:quantitative_results} summarizes quantitative proxies, while Table~\ref{tab:glm-fixed-effects} reports behavioral outcomes from our analyst study. All formulations are evaluated under matched data, models, and computational budgets. Fixed formulations use zero or mean baselines, while the remainder formulations rely on background resampling; conditional variants use proximity-weighted sampling, and counterfactual formulations employ optimization-based generation using DiCE~\cite{dice}.

\begin{table*}[t]
\centering
\caption{Average quantitative results across scale-invariant and scale-dependent metrics (bootstrapped standard deviations). Scale-dependent metrics use relative Rank statistics. \textbf{Bold} indicates the best outcome per metric.}
\setlength{\tabcolsep}{3.8pt}
\renewcommand{\arraystretch}{1.2}
\begin{tabular}{lccccccccc}
\toprule
\multirow{2}{*}{\textbf{Value F.}} & \multicolumn{4}{c}{\textbf{Rank Scale Dependent Metrics}} & \multicolumn{5}{c}{\textbf{Scale Invariant Metrics}} \\
\cmidrule(r){2-5} \cmidrule(l){6-10}
 & \textbf{Sparsity} & \textbf{Sensitivity} & \textbf{Contrast.} & \textbf{Attr. Error} & \textbf{Del. AUC} & \textbf{Ins. AUC} & \textbf{Recall@1} & \textbf{Recall@3} & \textbf{Recall@5} \\
\midrule
Base. Zero & 3.97 $\pm$ 2.74 & 4.10 $\pm$ 2.86 & 4.40 $\pm$ 2.99 & 3.41 $\pm$ 2.49 & \textbf{0.11 $\pm$ 0.08} & 0.88 $\pm$ 0.20 & \textbf{0.90 $\pm$ 0.07} & \textbf{0.90 $\pm$ 0.06} & \textbf{0.91 $\pm$ 0.05} \\
Base. Mean & 4.39 $\pm$ 2.39 & 5.21 $\pm$ 1.84 & 4.65 $\pm$ 1.56 & 3.46 $\pm$ 2.15 & 0.19 $\pm$ 0.13 & 0.83 $\pm$ 0.13 & 0.83 $\pm$ 0.11 & 0.85 $\pm$ 0.06 & 0.87 $\pm$ 0.07 \\
Uniform & 5.25 $\pm$ 2.30 & \textbf{2.81 $\pm$ 1.73} & 6.09 $\pm$ 2.20 & 5.00 $\pm$ 1.88 & 0.15 $\pm$ 0.24 & 0.76 $\pm$ 0.14 & 0.86 $\pm$ 0.10 & 0.85 $\pm$ 0.06 & 0.86 $\pm$ 0.06 \\
Marginal & 4.40 $\pm$ 1.34 & 3.66 $\pm$ 1.57 & 4.14 $\pm$ 1.93 & 3.57 $\pm$ 1.17 & 0.16 $\pm$ 0.10 & 0.75 $\pm$ 0.09 & 0.80 $\pm$ 0.12 & 0.84 $\pm$ 0.07 & 0.86 $\pm$ 0.08 \\
Joint-Marg. & 4.98 $\pm$ 1.95 & 3.51 $\pm$ 1.50 & 5.38 $\pm$ 1.22 & \textbf{3.15 $\pm$ 1.28} & 0.16 $\pm$ 0.10 & 0.75 $\pm$ 0.09 & 0.82 $\pm$ 0.08 & 0.82 $\pm$ 0.07 & 0.84 $\pm$ 0.07 \\
Conditional & 5.68 $\pm$ 1.65 & 4.73 $\pm$ 1.66 & 3.92 $\pm$ 1.09 & 6.24 $\pm$ 1.52 & 0.16 $\pm$ 0.11 & 0.77 $\pm$ 0.09 & 0.63 $\pm$ 0.24 & 0.72 $\pm$ 0.10 & 0.78 $\pm$ 0.09 \\
Counterf. & 5.17 $\pm$ 2.23 & 5.21 $\pm$ 2.66 & 5.94 $\pm$ 1.97 & 7.46 $\pm$ 0.86 & 0.20 $\pm$ 0.08 & 0.77 $\pm$ 0.09 & 0.53 $\pm$ 0.10 & 0.63 $\pm$ 0.10 & 0.70 $\pm$ 0.13 \\
Filt. Cond. & \textbf{2.15 $\pm$ 1.31} & 6.77 $\pm$ 1.42 & \textbf{1.47 $\pm$ 0.69} & 3.73 $\pm$ 2.00 & 0.18 $\pm$ 0.08 & \textbf{0.95 $\pm$ 0.05} & 0.90 $\pm$ 0.08 & 0.88 $\pm$ 0.06 & 0.89 $\pm$ 0.06 \\
\cmidrule(r){1-1} \cmidrule(r){2-6} \cmidrule(l){7-10}
\noalign{\vskip -1pt}
& \multicolumn{5}{c}{\small \textbf{Lower is better} ($\downarrow$)} & \multicolumn{4}{c}{\small \textbf{Higher is better} ($\uparrow$)} \\
\noalign{\vskip -1pt}
\bottomrule
\end{tabular}
\label{tab:quantitative_results}
\end{table*}

\subsection{Quantitative Benchmarks}
\label{sec:quantbenchmarks}

The benchmarks in Table~\ref{tab:quantitative_results} reveal structured trade-offs across the Shapley landscape. No single formulation dominates across all metrics. High \textit{sparsity} and \textit{contrastivity} improve insertion AUC, reflecting selective attributions, but increase \textit{sensitivity} to small input perturbations. Conversely, low-sensitivity formulations are stable and perform better on deletion tasks, but produce less selective explanations.

Across formulations, amortization achieves high fidelity to the KernelSHAP \textit{ground truth} reference, achieving low error and high Recall@$k$ for top-ranked features. Thus, observed properties and differences reflect semantic formulation choices rather than approximation artifacts. Also, cross-alignment analysis in Figure~\ref{fig:teaser} shows that formulations cluster into \textit{families}. Key trends include:
\begin{itemize}
    \item \textbf{Fixed baselines:} The Zero baseline yields the strongest Deletion AUC and Recall@$k$, but remains unaligned with empirical formulations. While the Uniform variant trades stability for reduced sparsity and contrastivity. The Mean baseline performs poorly across both both stability and selectivity metrics.
    \item \textbf{Empirical formulations:} Marginal and Joint-Marginal variants show mutual agreement and the most balanced performance, i.e. low sensitivity, moderate sparsity and contrastivity. Conditional Shapley departs from this pattern, producing dense, sensitive attributions that reflect \textit{data correlations rather than model behavior}~\cite{chen2020true}.    
    \item \textbf{Counterfactual formulations:} The Filtered Conditional variant maximizes sparsity, contrastivity, and Insertion AUC, but suffers from instability under perturbation. Fully counterfactual explanations are dense and unstable.
\end{itemize}

\begin{table*}[t]
    \centering
    \caption{Fixed effects on human outcomes. Decision time is reported as multiplicative effects; Accuracy, Clarity, and Confidence as Odds Ratios. Standard errors retrieved via the Delta Method. Bold values indicate significance ($p<0.05$).}
    \label{tab:glm-fixed-effects}
    \renewcommand{\arraystretch}{1.2}
    \setlength{\tabcolsep}{12.8pt}
    \begin{tabular}{lcccccc}
        \toprule
        \multirow{2}{*}{\textbf{Value Func.}} & 
        \multicolumn{3}{c}{\textbf{Decision Time {\small ($\downarrow$)}}} & 
        \multicolumn{3}{c}{\textbf{Odds Ratio {\small ($\uparrow$)}}} \\
        \cmidrule(lr){2-4} \cmidrule(lr){5-7}
        & \textbf{p2.5} & \textbf{p50} & \textbf{p97.5} & \textbf{Accuracy} & \textbf{Clarity} & \textbf{Confidence} \\
        \midrule

        Base. Zero  & \textbf{1.29 $\pm$ 0.09}  &\textbf{1.16 $\pm$ 0.05}   & 1.03 $\pm$ 0.07 & 1.04 $\pm$ 0.20 & \textbf{0.57 $\pm$ 0.06}  & 1.20 $\pm$ 0.15           \\
        Base. Mean  & 0.99 $\pm$ 0.07           & 1.03 $\pm$ 0.04           & 0.88 $\pm$ 0.06 & 1.05 $\pm$ 0.21 & 1.17 $\pm$ 0.14           & \textbf{1.49 $\pm$ 0.19}  \\
        Uniform     & 1.11 $\pm$ 0.08           & \textbf{1.11 $\pm$ 0.05}  & 1.02 $\pm$ 0.08 & 0.84 $\pm$ 0.16 & \textbf{0.79 $\pm$ 0.09}  & 1.02 $\pm$ 0.13           \\
        Marginal    & 0.97 $\pm$ 0.07           & 1.01 $\pm$ 0.04           & 1.08 $\pm$ 0.08 & 0.76 $\pm$ 0.14 & 1.18 $\pm$ 0.14           & 1.11 $\pm$ 0.14           \\
        Joint-Marg. & \textbf{1.17 $\pm$ 0.08}  & 1.06 $\pm$ 0.05           & 0.99 $\pm$ 0.07 & 1.19 $\pm$ 0.23 & \textbf{1.52 $\pm$ 0.19}  & 1.20 $\pm$ 0.16           \\
        Conditional & 0.98 $\pm$ 0.07           & 1.05 $\pm$ 0.04           & 0.94 $\pm$ 0.07 & 0.99 $\pm$ 0.19 & 1.23 $\pm$ 0.14           & \textbf{1.58 $\pm$ 0.20}  \\
        Counterf.   & 1.08 $\pm$ 0.07           & 1.00 $\pm$ 0.04           & 1.15 $\pm$ 0.08 & 1.02 $\pm$ 0.20 & 1.25 $\pm$ 0.16           & \textbf{1.55 $\pm$ 0.20}  \\
        Filt. Cond. & 1.03 $\pm$ 0.07           & 1.04 $\pm$ 0.05           & 0.98 $\pm$ 0.07 & 0.79 $\pm$ 0.15 & \textbf{0.68 $\pm$ 0.08}  & \textbf{1.30 $\pm$ 0.16}  \\

        \midrule

        Log Count & \textbf{0.84 $\pm$ 0.02} & \textbf{0.79 $\pm$ 0.01} & \textbf{0.80 $\pm$ 0.02} & 0.93 $\pm$ 0.06 & 1.01 $\pm$ 0.07 & 0.94 $\pm$ 0.05 \\
        Model Entropy & \textbf{1.25 $\pm$ 0.03} & \textbf{1.23 $\pm$ 0.02} & \textbf{1.07 $\pm$ 0.03} & 0.95 $\pm$ 0.08  & \textbf{0.68 $\pm$ 0.05} & \textbf{0.44 $\pm$ 0.05} \\
        Score Error & 1.00 $\pm$ 0.02  & 1.00 $\pm$ 0.02  & 0.99 $\pm$ 0.02 & \textbf{0.19 $\pm$ 0.02} & \textbf{0.86 $\pm$ 0.05} & \textbf{0.80 $\pm$ 0.03} \\
        Professional Analyst & \textbf{0.85 $\pm$ 0.05} & \textbf{0.88 $\pm$ 0.04} & 1.10 $\pm$ 0.07 & 0.94 $\pm$ 0.17   & 1.08 $\pm$ 0.17 & 1.10 $\pm$ 0.13 \\

        \bottomrule
    \end{tabular}
\end{table*}

\subsection{Human-in-the-Loop Utility}
\label{sec:qualitbenchmarks}

\begin{figure}[b]
    \includegraphics[width=0.47\textwidth]{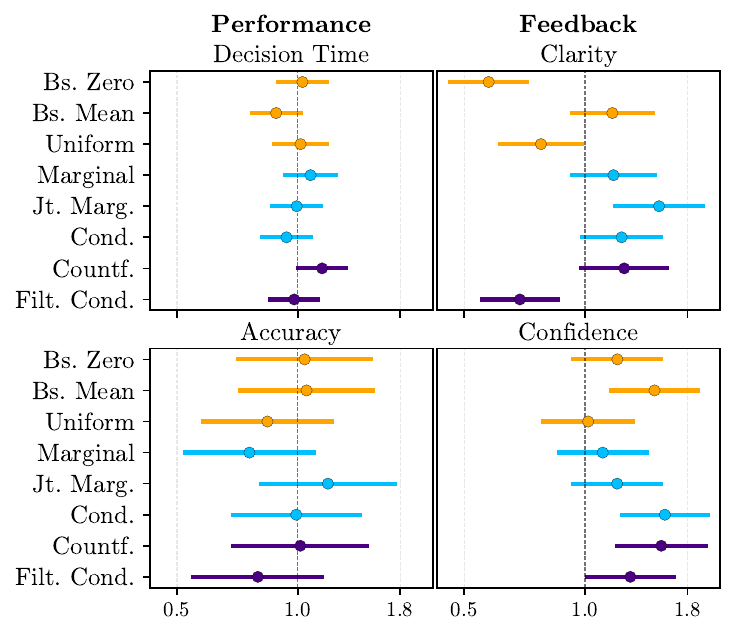}
    \caption{Effect sizes on human outcomes, with $95\%$ confidence intervals, across accuracy, decision time ($p_{97.5}$), confidence and clarity. X-axis is in log-scale.}
    \label{fig:qualitative_metrics}
    \Description{}
\end{figure}

Analyst behavior results in Table~\ref{tab:glm-fixed-effects} and Figure~\ref{fig:qualitative_metrics} reveal a critical asymmetry: formulation choice has no meaningful impact on objective performance (decision accuracy, review time) but strongly influences subjective outcomes (clarity, confidence). Explanations primarily shape how analysts \emph{interpret} and \emph{commit} to decisions rather than improving decision quality. Across formulations:
\begin{itemize}
    \item \textbf{Objective performance:} No formulation reliably improves accuracy or complex case (p97.5) decision time. Effects are driven by case-level factors such as model uncertainty, error or analyst exposure. Coupled with increased confidence, it highlights a severe risk of \textbf{automation bias}.
    \item \textbf{Perceived clarity} is sensitive to formulation choice. Joint-Marginal explanations significantly improve metrics, while the Zero baseline, Uniform, and Filtered Conditional formulations are consistently rated as confusing.
    \item \textbf{Confidence} shows strong explainer effects. Conditional, Counterfactual, and Mean formulations substantially inflate analyst confidence relative to no explanation.
\end{itemize}
Ultimately, clarity and confidence remain constrained by core signals of case difficulty, i.e. model uncertainty and prediction error. Explanations modulate analyst perception, but do not override task difficulty, aligning with recent literature on XAI and complementary performance~\cite{poursabzi2021manipulating, sivaraman2023ignore, bansal2021does}.

\subsection{Alignment: Proxies vs. Human Outcomes}
\label{sec:synthesis}

\begin{figure}[b]
    \includegraphics[width=0.48\textwidth]{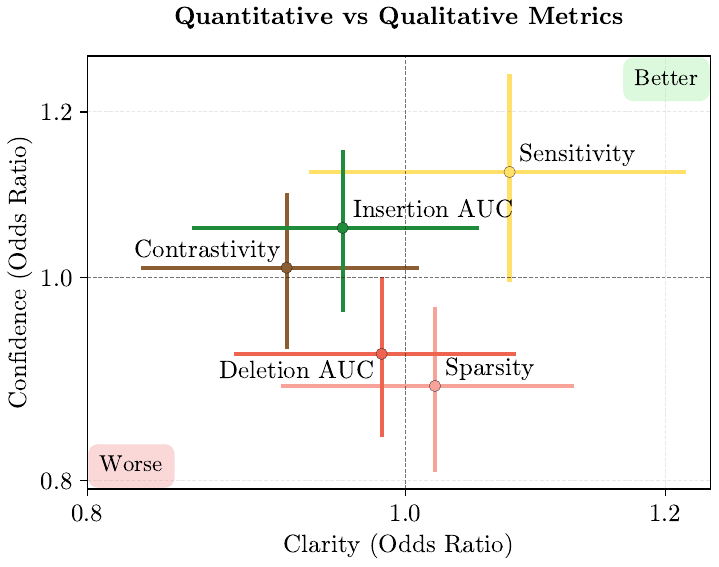}
    \caption{Effect of quantitative explanation metrics on clarity and decision confidence.}
    \label{fig:synthesis}
    \Description{}
\end{figure}

Finally, we audit the alignment between quantitative proxies and human outcomes by using functional properties as predictors for \textit{clarity} and \textit{confidence} (Figure~\ref{fig:synthesis}). In all cases, we control for case difficulty, prediction error, analyst experience, and expertise. 

Results reveal a \textbf{decoupling} between quantitative metrics and perceived utility: standard criteria do not show positive associations with clarity or confidence. Notably, \textit{sparsity} is \textbf{negatively} associated with confidence, contradicting the assumption that low-dimensional explanations enhance intelligibility~\cite{canha2025functionally}. Thus, widely used benchmarks capture structural properties but fail to predict human perception or use. While valuable for debugging, they are \textbf{not substitutes for behavioral evaluation} in operational settings.

\section{Conclusion}
\label{sec:discussion}

This paper has conducted a unified large-scale empirical study and audit of Shapley value formulations, testing the alignment between functional quantitative proxies and human utility in high-stakes risk workflows. By evaluating eight semantic variants under strictly matched data, model, interfaces, and computational conditions, we proved fundamental \textbf{decoupling} in XAI evaluation: standard quantitative benchmarks capture structural properties of explanations but fail to predict how they are perceived or used by final users. Furthermore, while no Shapley formulation improved objective accuracy, several significantly inflated analyst confidence. This uncovers a systemic risk of \textbf{automation bias} where explanations encourage over-reliance without providing cognitive gains.

\paragraph{Evidence-Based Guidance}
The results have direct implications for both research and practice. Based on our audit, we offer the following guidance for the deployment and evaluation of XAI:
\begin{itemize}
    \item \textbf{Decouple Algorithm from User:} Practitioners must treat proxy metrics (e.g., deletion AUC, sensitivity) as debugging tools for model-faithfulness, not as proxies for human interpretability or trust.
    \item System designers should \textbf{prioritize empirical formulations} for clarity (Marginal, Joint-Marginal, Conditional), while remaining wary of \emph{Conditional} or \emph{Counterfactual} variants that inflate confidence without commensurate gains in accuracy.
\end{itemize}

\paragraph{Limitations and Scope}
Our audit focused on low-latency, tabular risk models as a backbone to financial, e-commerce or medical decision-making~\cite{pmlr-v235-van-breugel24a}. However, results may differ in vision or language domains where feature semantics follow different dynamics. Additionally, our experiments were conducted in controlled settings and cannot capture longer-term effects such as learning, adaptation, or changes in institutional decision norms.

Ultimately, our results motivate a shift in the XAI community toward behaviorally grounded evaluation; until metrics are proven to align with human outcomes, they cannot serve as substitutes for rigorous user studies in high-stakes production systems.

\bibliographystyle{ACM-Reference-Format}
\bibliography{references}

@incollection{shapley1953value,
  title={A Value for n-Person Games},
  author={Shapley, Lloyd S},
  booktitle={Contributions to the Theory of Games II},
  _editor={Kuhn, Harold W. and Tucker, Albert W.},
  pages={307--317},
  year={1953},
  publisher={Princeton University Press},
  address={Princeton},
  _doi={10.1017/CBO9780511528446.003}
}

@inproceedings{lundberg2017unified,
  author={Lundberg, Scott M. and Lee, Su-In},
  title={A unified approach to interpreting model predictions},
  year={2017},
  publisher={Curran Associates Inc.},
  _address={Red Hook, NY, USA},
  booktitle={Proceedings of the 31st International Conference on Neural Information Processing Systems},
  pages={4768–4777},
  _url={https://proceedings.neurips.cc/paper_files/paper/2017/file/8a20a8621978632d76c43dfd28b67767-Paper.pdf}
}

@misc{lundberg2018consistent,
  title={Consistent Individualized Feature Attribution for Tree Ensembles}, 
  author={Scott M. Lundberg and Gabriel G. Erion and Su-In Lee},
  year={2019},
  eprint={1802.03888},
  archivePrefix={arXiv},
  primaryClass={cs.LG},
  _url={https://arxiv.org/abs/1802.03888}, 
}

@inproceedings{jethani2022fastshap,
  title={Fast{SHAP}: Real-Time Shapley Value Estimation},
  author={Neil Jethani and Mukund Sudarshan and Ian Connick Covert and Su-In Lee and Rajesh Ranganath},
  booktitle={International Conference on Learning Representations},
  year={2022},
  _url={https://openreview.net/forum?id=Zq2G_VTV53T},
  numpages={}
}

@InProceedings{sundararajan2020many,
  title={The Many Shapley Values for Model Explanation},
  author={Sundararajan, Mukund and Najmi, Amir},
  booktitle={Proceedings of the 37th International Conference on Machine Learning},
  pages={9269--9278},
  year={2020},
  _editor={III, Hal Daumé and Singh, Aarti},
  volume={119},
  series={Proceedings of Machine Learning Research},
  publisher={PMLR},
  _pdf={http://proceedings.mlr.press/v119/sundararajan20b/sundararajan20b.pdf},
  _url={https://proceedings.mlr.press/v119/sundararajan20b.html},
}

@inproceedings{bhatt2020explainable,
  author={Bhatt, Umang and Xiang, Alice and Sharma, Shubham and Weller, Adrian and Taly, Ankur and Jia, Yunhan and Ghosh, Joydeep and Puri, Ruchir and Moura, Jos\'{e} M. F. and Eckersley, Peter},
  title={Explainable machine learning in deployment},
  year={2020},
  _isbn={9781450369367},
  publisher={Association for Computing Machinery},
  _address={New York, NY, USA},
  _url={https://doi.org/10.1145/3351095.3375624},
  _doi={10.1145/3351095.3375624},
  booktitle={Proceedings of the 2020 Conference on Fairness, Accountability, and Transparency},
  pages={648–657},
  _numpages={10},
  _keywords={deployed systems, explainability, machine learning, qualitative study, transparency},
  _location={Barcelona, Spain},
  _series={FAT* '20}
}

@inproceedings{poursabzi2021manipulating,
  title={Manipulating and measuring model interpretability},
  author={Forough Poursabzi-Sangdeh and Daniel G Goldstein and Jake M Hofman and Jennifer Wortman Vaughan and Hanna Wallach},
  booktitle={Proceedings of the 2021 CHI conference on human factors in computing systems},
  pages={1--52},
  year={2021}
}

@article{khan2025model,
  author={Khan, Farhina Sardar and Mazhar, Syed Shahid and Mazhar, Kashif and A. AlSaleh, Dhoha and Mazhar, Amir},
  title={Model-agnostic explainable artificial intelligence methods in finance: a systematic review, recent developments, limitations, challenges and future directions},
  journal={Artificial Intelligence Review},
  year={2025},
  volume={58},
  number={8},
  pages={232},
  _issn={1573-7462},
  _doi={10.1007/s10462-025-11215-9},
  _url={https://doi.org/10.1007/s10462-025-11215-9}
}

@article{mienye2024survey,
  title={A survey of explainable artificial intelligence in healthcare: Concepts, applications, and challenges},
  journal={Informatics in Medicine Unlocked},
  volume={51},
  pages={101587},
  year={2024},
  _issn={2352-9148},
  _doi={https://doi.org/10.1016/j.imu.2024.101587},
  _url={https://www.sciencedirect.com/science/article/pii/S2352914824001448},
  author={Ibomoiye Domor Mienye and George Obaido and Nobert Jere and Ebikella Mienye and Kehinde Aruleba and Ikiomoye Douglas Emmanuel and Blessing Ogbuokiri},
  _keywords={AI, Bias, Ethics, Fairness, Healthcare, Machine learning},
}

@article{bucker2022transparency,
  author={Michael Bücker and Gero Szepannek and Alicja Gosiewska and Przemyslaw Biecek},
  title={Transparency, auditability, and explainability of machine learning models in credit scoring},
  journal={Journal of the Operational Research Society},
  volume={73},
  number={1},
  pages={70--90},
  year={2022},
  publisher={Taylor \& Francis},
  _doi={10.1080/01605682.2021.1922098},
  _url={https://doi.org/10.1080/01605682.2021.1922098}
}

@article{yang2021fast,
  title={Fast TreeSHAP: Accelerating SHAP Value Computation for Trees},
  author={Jilei Yang},
  journal={ArXiv},
  year={2021},
  volume={abs/2109.09847},
  pages={},
  _url={https://api.semanticscholar.org/CorpusID:237581337}
}

@inproceedings{bifet2022linear,
   author={Yu, Peng and Bifet, Albert and Read, Jesse and Xu, Chao},
   booktitle={Advances in Neural Information Processing Systems},
   _editor={S. Koyejo and S. Mohamed and A. Agarwal and D. Belgrave and K. Cho and A. Oh},
   pages={25818--25828},
   publisher={Curran Associates, Inc.},
   title={Linear tree shap},
   _url={https://proceedings.neurips.cc/paper_files/paper/2022/file/a5a3b1ef79520b7cd122d888673a3ebc-Paper-Conference.pdf},
   _address={Red Hook, NY, USA},
   volume={35},
   year={2022}
}

@inproceedings{muschalik2024beyond,
  author={Muschalik, Maximilian and Fumagalli, Fabian and Hammer, Barbara and H\"{u}llermeier, Eyke},
  title={Beyond TreeSHAP: efficient computation of any-order shapley interactions for tree ensembles},
  year={2024},
  _isbn={978-1-57735-887-9},
  publisher={AAAI Press},
  _url={https://doi.org/10.1609/aaai.v38i13.29352},
  _doi={10.1609/aaai.v38i13.29352},
  booktitle={Proceedings of the Thirty-Eighth AAAI Conference on Artificial Intelligence},
  articleno={1604},
  numpages={9},
  series={AAAI'24}
}

@article{rudin2019stop,
  author={Rudin, Cynthia},
  title={Stop explaining black box machine learning models for high stakes decisions and use interpretable models instead},
  journal={Nature Machine Intelligence},
  year={2019},
  volume={1},
  number={5},
  pages={206-215},
  _issn={2522-5839},
  _doi={10.1038/s42256-019-0048-x},
  _url={https://doi.org/10.1038/s42256-019-0048-x}
}

@inproceedings{albini2022counterfactual,
  author={Albini, Emanuele and Long, Jason and Dervovic, Danial and Magazzeni, Daniele},
  title={Counterfactual Shapley Additive Explanations},
  year={2022},
  _isbn={9781450393522},
  publisher={Association for Computing Machinery},
  _address={New York, NY, USA},
  _doi={10.1145/3531146.3533168},
  booktitle={Proceedings of the 2022 ACM Conference on Fairness, Accountability, and Transparency},
  pages={1054–1070},
  numpages={17},
  _location={Seoul, Republic of Korea},
  series={FAccT '22}
}

@inproceedings{taufiq2023manifold,
  title={Manifold Restricted Interventional Shapley Values},
  author={Taufiq, Muhammad Faaiz and Bl\"obaum, Patrick and Minorics, Lenon},
  booktitle={Proceedings of The 26th International Conference on Artificial Intelligence and Statistics},
  pages={5079--5106},
  year={2023},
  _editor={Ruiz, Francisco and Dy, Jennifer and van de Meent, Jan-Willem},
  volume={206},
  series={Proceedings of Machine Learning Research},
  publisher={PMLR},
  _pdf={https://proceedings.mlr.press/v206/taufiq23a/taufiq23a.pdf},
  _url={https://proceedings.mlr.press/v206/taufiq23a.html},
}

@InProceedings{meyer2023training,
  title={Training Deep Surrogate Models with Large Scale Online Learning},
  author={Meyer, Lucas Thibaut and Schouler, Marc and Caulk, Robert Alexander and Ribes, Alejandro and Raffin, Bruno},
  booktitle={Proceedings of the 40th International Conference on Machine Learning},
  pages={24614--24630},
  year={2023},
  volume={202},
  publisher={PMLR},
  _pdf={https://proceedings.mlr.press/v202/meyer23b/meyer23b.pdf},
  _url={https://proceedings.mlr.press/v202/meyer23b.html},
}

@inproceedings{dyer2024interventionally,
 author={Dyer, Joel and Bishop, Nicholas and Felekis, Yorgos and Zennaro, Fabio Massimo and Calinescu, Anisoara and Damoulas, Theodoros and Wooldridge, Michael},
 booktitle={Advances in Neural Information Processing Systems},
 _doi={10.52202/079017-0686},
 pages={21814--21841},
 publisher={Curran Associates, Inc.},
 _address={Red Hook, NY, USA},
 title={Interventionally Consistent Surrogates for Complex Simulation Models},
 volume={37},
 year={2024}
}

@inproceedings{goldwasser2024stabilizing,
  author={Goldwasser, Jeremy and Hooker, Giles},
  title={Stabilizing Estimates of Shapley Values with Control Variates},
  booktitle={Explainable Artificial Intelligence},
  year={2024},
  publisher={Springer Nature Switzerland},
  _address={Cham},
  pages={416--439},
  _isbn={978-3-031-63797-1},
  _doi={https://doi.org/10.1007/978-3-031-63797-1_21}
}

@article{zern2023interventional,
  title={Interventional SHAP Values and Interaction Values for Piecewise Linear Regression Trees},
  author={Zern, Artjom and Broelemann, Klaus and Kasneci, Gjergji},
  journal={Proceedings of the AAAI Conference on Artificial Intelligence},
  year={2023},
  volume={37},
  pages={11164-11173},
  _url={https://ojs.aaai.org/index.php/AAAI/article/view/26322},
  _doi={10.1609/aaai.v37i9.26322},
  number={9},
}

@article{aas2021explaining,
  title={Explaining individual predictions when features are dependent: More accurate approximations to Shapley values},
  journal={Artificial Intelligence},
  volume={298},
  pages={103502},
  year={2021},
  _issn={0004-3702},
  _doi={https://doi.org/10.1016/j.artint.2021.103502},
  _url={https://www.sciencedirect.com/science/article/pii/S0004370221000539},
  author={Kjersti Aas and Martin Jullum and Anders Løland},
  _keywords={Feature attribution, Shapley values, Kernel SHAP, Dependence}
}

@inproceedings{covert2024stochastic,
  author={Covert, Ian and Kim, Chanwoo and Lee, Su-In and Zou, James and Hashimoto, Tatsunori},
  title={Stochastic amortization: a unified approach to accelerate feature and data attribution},
  year={2024},
  _isbn={9798331314385},
  publisher={Curran Associates Inc.},
  _address={Red Hook, NY, USA},
  booktitle={Proceedings of the 38th International Conference on Neural Information Processing Systems},
  articleno={143},
  _doi={https://doi.org/10.52202/079017-0143},
  numpages={50},
  _location={Vancouver, BC, Canada},
  series={NIPS '24}
}

@inproceedings{covert2021improving,
  title={Improving KernelSHAP: Practical Shapley Value Estimation Using Linear Regression},
  author={Covert, Ian and Lee, Su-In},
  booktitle={Proceedings of The 24th International Conference on Artificial Intelligence and Statistics},
  pages={3457--3465},
  year={2021},
  volume={130},
  series={Proceedings of Machine Learning Research},
  publisher={PMLR},
  _address={Indio, CA, USA},
  _pdf={http://proceedings.mlr.press/v130/covert21a/covert21a.pdf},
  _url={https://proceedings.mlr.press/v130/covert21a.html}
}

@article{chen2023algorithms,
  author={Chen, Hugh and Covert, Ian C. and Lundberg, Scott M. and Lee, Su-In},
  title={Algorithms to estimate Shapley value feature attributions},
  journal={Nature Machine Intelligence},
  year={2023},
  volume={5},
  number={6},
  pages={590-601},
  _issn={2522-5839},
  _doi={10.1038/s42256-023-00657-x},
  _url={https://doi.org/10.1038/s42256-023-00657-x}
}

@InProceedings{olsen2024improving,
author={Olsen, Lars Henry Berge and Jullum, Martin},
_editor={Guidotti, Riccardo and Schmid, Ute and Longo, Luca},
title={Improving the Weighting Strategy in KernelSHAP},
booktitle={Explainable Artificial Intelligence},
year={2026},
publisher={Springer Nature Switzerland},
_address={Cham},
pages={194--218},
_isbn={978-3-032-08324-1},
_doi={10.1007/978-3-032-08324-1_9},
_url={https://doi.org/10.1007/978-3-032-08324-1_9}
}

@InProceedings{janzing2020feature,
  title={Feature relevance quantification in explainable AI: A causal problem},
  author={Janzing, Dominik and Minorics, Lenon and Bloebaum, Patrick},
  booktitle={Proceedings of the Twenty Third International Conference on Artificial Intelligence and Statistics},
  pages={2907--2916},
  year={2020},
  volume={108},
  series={Proceedings of Machine Learning Research},
  publisher={PMLR},
  _pdf={http://proceedings.mlr.press/v108/janzing20a/janzing20a.pdf},
  _url={https://proceedings.mlr.press/v108/janzing20a.html},
}

@inproceedings{heskes2020causal,
  author={Heskes, Tom and Sijben, Evi and Bucur, Ioan Gabriel and Claassen, Tom},
  booktitle={Advances in Neural Information Processing Systems},
  pages={4778--4789},
  publisher={Curran Associates, Inc.},
  title={Causal Shapley Values: Exploiting Causal Knowledge to Explain Individual Predictions of Complex Models},
  volume={33},
  year={2020},
  _address={Red Hook, NY, USA},
  _url={https://proceedings.neurips.cc/paper_files/paper/2020/file/32e54441e6382a7fbacbbbaf3c450059-Paper.pdf}
}

@article{covert2021explaining,
  author={Ian Covert and Scott Lundberg and Su-In Lee},
  title={Explaining by Removing: A Unified Framework for Model Explanation},
  journal={Journal of Machine Learning Research},
  year={2021},
  volume={22},
  number={209},
  pages={1--90},
  _url={http://jmlr.org/papers/v22/20-1316.html}
}

@inproceedings{merrick2020explanation,
  title={The Explanation Game: Explaining Machine Learning Models Using Shapley Values},
  author={Merrick, Luke and Taly, Ankur},
  booktitle={International Cross-Domain Conference for Machine Learning and Knowledge Extraction},
  year={2020},
  publisher={Springer International Publishing},
  _address={Cham},
  pages={17--38},
  _doi={https://doi.org/10.1007/978-3-030-57321-8_2}
}

@inproceedings{datta2016algorithmic,
  author={Datta, Anupam and Sen, Shayak and Zick, Yair},
  booktitle={2016 IEEE Symposium on Security and Privacy (SP)},
  title={Algorithmic Transparency via Quantitative Input Influence: Theory and Experiments with Learning Systems},
  year={2016},
  volume={},
  _ISSN={2375-1207},
  pages={598-617},
  _doi={10.1109/SP.2016.42},
  publisher={IEEE Computer Society},
  _address={Los Alamitos, CA, USA},
}

@article{strumbelj2010efficient,
  author={Erik {\v{S}}trumbelj and Igor Kononenko},
  title={An Efficient Explanation of Individual Classifications using Game Theory},
  journal={Journal of Machine Learning Research},
  year={2010},
  volume={11},
  number={1},
  pages={1--18},
  _url={http://jmlr.org/papers/v11/strumbelj10a.html}
}

@inproceedings{karimi2021algorithmic,
  author={Karimi, Amir-Hossein and Sch\"{o}lkopf, Bernhard and Valera, Isabel},
  title={Algorithmic Recourse: from Counterfactual Explanations to Interventions},
  year={2021},
  _isbn={9781450383097},
  publisher={Association for Computing Machinery},
  _address={New York, NY, USA},
  _doi={10.1145/3442188.3445899},
  booktitle={Proceedings of the 2021 ACM Conference on Fairness, Accountability, and Transparency},
  pages={353–362},
  numpages={10},
  _keywords={algorithmic recourse, causal inference, consequential recommendations, contrastive explanations, counterfactual explanations, explainable artificial intelligence, minimal interventions},
  _location={Virtual Event, Canada},
  series={FAccT '21}
}

@inproceedings{molnar2020interpretable,
  title={Interpretable Machine Learning -- A Brief History, State-of-the-Art and Challenges},
  author={Molnar, Christoph and Casalicchio, Giuseppe and Bischl, Bernd},
  booktitle={ECML PKDD 2020 Workshops},
  year={2020},
  publisher={Springer International Publishing},
  _address={Cham},
  pages={417--431},
  _doi={https://doi.org/10.1007/978-3-030-65965-3_28}
}

@article{canha2025functionally,
  author={Canha, Dulce and Kubler, Sylvain and Fr\"{a}mling, Kary and Fagherazzi, Guy},
  title={A Functionally-Grounded Benchmark Framework for XAI Methods: Insights and Foundations from a Systematic Literature Review},
  year={2025},
  issue_date={December 2025},
  publisher={Association for Computing Machinery},
  _address={New York, NY, USA},
  volume={57},
  number={12},
  _issn={0360-0300},
  _url={https://doi.org/10.1145/3737445},
  _doi={10.1145/3737445},
  journal={ACM Comput. Surv.},
  articleno={320},
  numpages={40},
  _keywords={Artificial intelligence, machine learning, eXplainable AI (XAI), transparency, interpretability, trustworthiness, responsible AI}
}

@article{ali2023explainable,
  title={Explainable Artificial Intelligence (XAI): What we know and what is left to attain Trustworthy Artificial Intelligence},
  journal={Information Fusion},
  volume={99},
  pages={101805},
  year={2023},
  _issn={1566-2535},
  _doi={https://doi.org/10.1016/j.inffus.2023.101805},
  _url={https://www.sciencedirect.com/science/article/pii/S1566253523001148},
  author={Sajid Ali and Tamer Abuhmed and Shaker El-Sappagh and Khan Muhammad and Jose M. Alonso-Moral and Roberto Confalonieri and Riccardo Guidotti and Javier {Del Ser} and Natalia Díaz-Rodríguez and Francisco Herrera},
  _keywords={Explainable Artificial Intelligence, Interpretable machine learning, Trustworthy AI, AI principles, Post-hoc explainability, XAI assessment, Data Fusion, Deep Learning},
}

@misc{doshi2017towards,
  title={Towards A Rigorous Science of Interpretable Machine Learning}, 
  author={Finale Doshi-Velez and Been Kim},
  year={2017},
  eprint={1702.08608},
  archivePrefix={arXiv},
  primaryClass={stat.ML},
}

@article{vilone2021notions,
title={Notions of explainability and evaluation approaches for explainable artificial intelligence},
journal={Information Fusion},
volume={76},
pages={89-106},
year={2021},
_issn={1566-2535},
_doi={https://doi.org/10.1016/j.inffus.2021.05.009},
_url={https://www.sciencedirect.com/science/article/pii/S1566253521001093},
author={Giulia Vilone and Luca Longo},
_keywords={Explainable artificial intelligence, Notions of explainability, Evaluation methods}
}

@article{saeed2023explainable,
  title={Explainable AI (XAI): A systematic meta-survey of current challenges and future opportunities},
  journal={Knowledge-Based Systems},
  volume={263},
  pages={110273},
  year={2023},
  _issn={0950-7051},
  _doi={https://doi.org/10.1016/j.knosys.2023.110273},
  _url={https://www.sciencedirect.com/science/article/pii/S0950705123000230},
  author={Waddah Saeed and Christian Omlin},
  _keywords={Explainable AI (XAI), Interpretable AI, Black-box, Machine learning, Deep learning, Meta-survey, Responsible AI},
}

@article{nauta2023anecdotal,
  author={Nauta, Meike and Trienes, Jan and Pathak, Shreyasi and Nguyen, Elisa and Peters, Michelle and Schmitt, Yasmin and Schl\"{o}tterer, J\"{o}rg and van Keulen, Maurice and Seifert, Christin},
  title={From Anecdotal Evidence to Quantitative Evaluation Methods: A Systematic Review on Evaluating Explainable AI},
  year={2023},
  publisher={Association for Computing Machinery},
  _address={New York, NY, USA},
  volume={55},
  number={13s},
  _issn={0360-0300},
  _url={https://doi.org/10.1145/3583558},
  _doi={10.1145/3583558},
  journal={ACM Comput. Surv.},
  _articleno={295},
  _keywords={XAI, explainable AI, quantitative evaluation methods, interpretability, explainability, evaluation, interpretable machine learning, Explainable artificial intelligence},
  numpages={}
}

@inproceedings{ding2021retiring,
  author={Ding, Frances and Hardt, Moritz and Miller, John and Schmidt, Ludwig},
  booktitle={Advances in Neural Information Processing Systems},
  pages={6478--6490},
  publisher={Curran Associates, Inc.},
  title={Retiring Adult: New Datasets for Fair Machine Learning},
  _address={Red Hook, NY, USA},
  volume={34},
  year={2021},
  _url={https://proceedings.neurips.cc/paper_files/paper/2021/file/32e54441e6382a7fbacbbbaf3c450059-Paper.pdf}
}

@InProceedings{ahmed2020review,
  author={Ahmed, Marzia and Kashem, Mohammod Abul and Rahman, Mostafijur and Khatun, Sabira},
  title={Review and Analysis of Risk Factor of Maternal Health in Remote Area Using the Internet of Things (IoT)},
  booktitle={InECCE2019},
  year={2020},
  publisher={Springer Singapore},
  _address={Singapore},
  pages={357--365},
  isbn={978-981-15-2317-5},
  _doi={https://doi.org/10.1007/978-981-15-2317-5_30}
}

@online{fico2025educational-analytics,
  author={{FICO}},
  title={FICO Expands Educational Analytics Challenge Program with Three New Historically Black Colleges and Universities to Educate Aspiring Data Scientists},
  year={2025},
  _url={https://www.fico.com/en/newsroom/fico-expands-educational-analytics-challenge-program-three-new-historically-black-colleges-and-universities-educate-aspiring-data-scientists},
  note={Accessed: 2026-01-02},
  organization={FICO}
}

@InProceedings{perez2022attribution,
  title={Attribution of predictive uncertainties in classification models},
  author={Perez, Iker and Skalski, Piotr and Barns-Graham, Alec and Wong, Jason and Sutton, David},
  booktitle={Proceedings of the Thirty-Eighth Conference on Uncertainty in Artificial Intelligence},
  pages={1582--1591},
  year={2022},
  _editor={Cussens, James and Zhang, Kun},
  volume={180},
  series={Proceedings of Machine Learning Research},
  publisher={PMLR},
  _pdf={https://proceedings.mlr.press/v180/perez22a/perez22a.pdf},
  _url={https://proceedings.mlr.press/v180/perez22a.html},
}

@inproceedings{ke2017lightgbm,
  author={Ke, Guolin and Meng, Qi and Finley, Thomas and Wang, Taifeng and Chen, Wei and Ma, Weidong and Ye, Qiwei and Liu, Tie-Yan},
  booktitle={Advances in Neural Information Processing Systems},
  pages={},
  publisher={Curran Associates, Inc.},
  title={LightGBM: A Highly Efficient Gradient Boosting Decision Tree},
  _url={https://proceedings.neurips.cc/paper_files/paper/2017/file/6449f44a102fde848669bdd9eb6b76fa-Paper.pdf},
  _address={Red Hook, NY, USA},
  volume={30},
  year={2017}
}

@inproceedings{jesus2021how,
  author={Jesus, S\'{e}rgio and Bel\'{e}m, Catarina and Balayan, Vladimir and Bento, Jo\~{a}o and Saleiro, Pedro and Bizarro, Pedro and Gama, Jo\~{a}o},
  title={How can I choose an explainer? An Application-grounded Evaluation of Post-hoc Explanations},
  year={2021},
  isbn={9781450383097},
  publisher={Association for Computing Machinery},
  _address={New York, NY, USA},
  _url={https://doi.org/10.1145/3442188.3445941},
  _doi={10.1145/3442188.3445941},
  booktitle={Proceedings of the 2021 ACM Conference on Fairness, Accountability, and Transparency},
  _pages={805–815},
  numpages={11},
  _keywords={Evaluation, Explainability, LIME, SHAP, User Study, XAI},
  _location={Virtual Event, Canada},
  _series={FAccT '21}
}

@article{amarasinghe2024importance,
  title={On the Importance of Application-Grounded Experimental Design for Evaluating Explainable ML Methods},
  volume={38},
  _url={https://ojs.aaai.org/index.php/AAAI/article/view/30082},
  _doi={10.1609/aaai.v38i19.30082}, 
  _number={19},
  journal={Proceedings of the AAAI Conference on Artificial Intelligence},
  author={Amarasinghe, Kasun and Rodolfa, Kit T. and Jesus, Sérgio and Chen, Valerie and Balayan, Vladimir and Saleiro, Pedro and Bizarro, Pedro and Talwalkar, Ameet and Ghani, Rayid},
  year={2024},
  pages={20921-20929}
}

@inproceedings{optuna,
  author={Akiba, Takuya and Sano, Shotaro and Yanase, Toshihiko and Ohta, Takeru and Koyama, Masanori},
  title={Optuna: A Next-generation Hyperparameter Optimization Framework},
  year={2019},
  isbn={9781450362016},
  publisher={Association for Computing Machinery},
  _address={New York, NY, USA},
  _url={https://doi.org/10.1145/3292500.3330701},
  _doi={10.1145/3292500.3330701},
  booktitle={Proceedings of the 25th ACM SIGKDD International Conference on Knowledge Discovery \& Data Mining},
  _pages={2623–2631},
  numpages={9},
  _keywords={Bayesian optimization, black-box optimization, hyperparameter optimization, machine learning system},
  _location={Anchorage, AK, USA},
  _series={KDD '19}
}

@article{aequitas,
  author={S{{\'e}}rgio Jesus and Pedro Saleiro and In{{\^e}}s Oliveira e Silva and Beatriz M. Jorge and Rita P. Ribeiro and Jo{{\~a}}o Gama and Pedro Bizarro and Rayid Ghani},
  title={Aequitas Flow: Streamlining Fair ML Experimentation},
  journal={Journal of Machine Learning Research},
  year={2024},
  volume={25},
  _number={354},
  pages={1--7},
  _url={http://jmlr.org/papers/v25/24-0677.html},
}

@inproceedings{dice,
  author={Mothilal, Ramaravind K. and Sharma, Amit and Tan, Chenhao},
  title={Explaining machine learning classifiers through diverse counterfactual explanations},
  year={2020},
  _isbn={9781450369367},
  publisher={Association for Computing Machinery},
  address={New York, NY, USA},
  _url={https://doi.org/10.1145/3351095.3372850},
  _doi={10.1145/3351095.3372850},
  booktitle={Proceedings of the 2020 Conference on Fairness, Accountability, and Transparency},
  _pages={607–617},
  numpages={11},
  _location={Barcelona, Spain},
  _series={FAT* '20}
}

@article{wysocki2023assessing,
  title={Assessing the communication gap between AI models and healthcare professionals: Explainability, utility and trust in AI-driven clinical decision-making},
  journal={Artificial Intelligence},
  volume={316},
  pages={103839},
  year={2023},
  _issn={0004-3702},
  _doi={https://doi.org/10.1016/j.artint.2022.103839},
  _url={https://www.sciencedirect.com/science/article/pii/S0004370222001795},
  author={Oskar Wysocki and Jessica Katharine Davies and Markel Vigo and Anne Caroline Armstrong and Dónal Landers and Rebecca Lee and André Freitas},
}

@book{pinheiro2000mixed,
  title={Mixed-effects models in S and S-PLUS},
  author={Pinheiro, Jos{\'e} C and Bates, Douglas M},
  year={2000},
  publisher={Springer},
  _doi={https://doi.org/10.1007/b98882}
}

@book{fitzmaurice2012applied,
  title={Applied longitudinal analysis},
  author={Fitzmaurice, Garrett M and Laird, Nan M and Ware, James H},
  year={2012},
  publisher={John Wiley \& Sons},
  _doi={https://doi.org/10.1002/9781119513469}
}

@article{nelder1972generalized,
  title={Generalized linear models},
  author={Nelder, John Ashworth and Wedderburn, Robert WM},
  journal={Journal of the Royal Statistical Society Series A: Statistics in Society},
  volume={135},
  _number={3},
  pages={370--384},
  year={1972},
  publisher={Oxford University Press},
  _doi={https://doi.org/10.2307/2344614}
}

@InProceedings{pmlr-v235-van-breugel24a,
  title = {Position: Why Tabular Foundation Models Should Be a Research Priority},
  author = {Van Breugel, Boris and Van Der Schaar, Mihaela},
  booktitle = {Proceedings of the 41st International Conference on Machine Learning},
  pages =  {48976--48993},
  year = {2024},
  volume = 	 {235},
  publisher =    {PMLR},
}

@inproceedings{sivaraman2023ignore,
  title={Ignore, trust, or negotiate: understanding clinician acceptance of AI-based treatment recommendations in health care},
  author={Sivaraman, Venkatesh and Bukowski, Leigh A and Levin, Joel and Kahn, Jeremy M and Perer, Adam},
  booktitle={Proceedings of the 2023 CHI Conference on Human Factors in Computing Systems},
  pages={1--18},
  year={2023}
}

@inproceedings{bansal2021does,
  title={Does the whole exceed its parts? the effect of ai explanations on complementary team performance},
  author={Bansal, Gagan and Wu, Tongshuang and Zhou, Joyce and Fok, Raymond and Nushi, Besmira and Kamar, Ece and Ribeiro, Marco Tulio and Weld, Daniel},
  booktitle={Proceedings of the 2021 CHI conference on human factors in computing systems},
  pages={1--16},
  year={2021}
}

@article{chen2020true,
  title={True to the model or true to the data?},
  author={Chen, Hugh and Janizek, Joseph D and Lundberg, Scott and Lee, Su-In},
  journal={arXiv preprint arXiv:2006.16234},
  year={2020}
}

@article{grinsztajn2022tree,
  title={Why do tree-based models still outperform deep learning on typical tabular data?},
  author={Grinsztajn, L{\'e}o and Oyallon, Edouard and Varoquaux, Ga{\"e}l},
  journal={Advances in neural information processing systems},
  volume={35},
  pages={507--520},
  year={2022}
}

@article{sahakyan2021explainable,
  title={Explainable artificial intelligence for tabular data: A survey},
  author={Sahakyan, Maria and Aung, Zeyar and Rahwan, Talal},
  journal={IEEE access},
  volume={9},
  pages={135392--135422},
  year={2021},
  publisher={IEEE}
}


\appendix

\section{Extended Overview of Shapley Value Formulations}
\label{app:shapleyformulations}

The definition of the value function $v_{\boldsymbol{x}}(\mathcal{S})$ hinges on how one represents feature \textbf{absence}. These formulations are not interchangeable; they encode fundamental assumptions about the data-generating process and the intended use of the explanation. Table~\ref{tab:shapdefinitions} summarizes the formulations considered in this work. Here, we extend their definitions with a discussion on use cases and limitations.

\begin{table*}[t!]
    \renewcommand{\arraystretch}{1.25}
    \setlength{\tabcolsep}{4pt}
    \centering
    \caption{Overview of key functional properties of explanation quality~\cite{taufiq2023manifold}, with example attributes.}
    \begin{tabular}{p{3.4cm}p{5.4cm}p{8.0cm}}
    \toprule
    \textbf{Property~\cite{taufiq2023manifold}} & \textbf{What it captures} & \textbf{Example quantitative / qualitative attributes} \\
    \midrule
    \textbf{Representativeness} & \textbf{What} explanation is provided. & Local vs. global, model portability or data coverage. \\
    \textbf{Structure} & \textbf{How} is the explanation provided. & Expressive power, graphical integrity, morphological clarity. \\
    \textbf{Selectivity} & \textbf{Size} of the explanation. & How concise. Sparsity and top-$k$ relevance. \\
    \textbf{Contrastivity} & \textbf{Relational} differences. & Sensitivity to counterfactuals, benchmarks or perturbations. \\
    \textbf{Interactivity} & User \textbf{control}. & Interaction options, adjustable parameters or interfaces. \\
    \textbf{Fidelity} & \textbf{Surrogate} consistency & Approximation accuracy and model assumptions. \\
    \textbf{Faithfulness} & How \textbf{reliable} for the black-box. & Causal correctness, feature removal and white-box checks. \\
    \textbf{Truthfulness} & Domain \textbf{alignment}. & Expert plausibility; bias exposure. \\
    \textbf{Stability} & Behavioural \textbf{consistency}. & Similarity across perturbations or resampling. \\
    \textbf{(Un)certainty} & How \textbf{transparent} is it? & Confidence disclosure, stochastic awareness. \\
    \textbf{Speed} & Computational \textbf{efficiency}. & Latency; throughput under fixed budgets. \\
    \bottomrule
    \end{tabular}

    \label{tab:criteria}
\end{table*}

\subsection{Fixed Baseline Shapley Values}

The \textit{fixed baseline} formulation replaces absent features with values from a predefined reference input $\boldsymbol{x}'$~\cite{sundararajan2020many, lundberg2017unified}. The background distribution is:
\begin{equation*}
\label{eq:baseline}
p(\cdot) = \delta(\boldsymbol{X}_{\mathcal{F} \setminus \mathcal{S}} = \boldsymbol{x}'_{\mathcal{F} \setminus \mathcal{S}}).
\end{equation*}
While computationally efficient (requiring only one model evaluation per coalition), this method is highly sensitive to the choice of $\boldsymbol{x}'$. As there is rarely a "neutral" reference point in tabular data~\cite{chen2023algorithms}, the choice of zero, mean, or median can fundamentally shift the attribution logic.

\subsection{Marginal (Interventional) Shapley Values}

In the \textit{marginal} formulation~\cite{sundararajan2020many, janzing2020feature}, absent features are sampled from their global empirical distribution:
\begin{equation*}
\label{eq:marginal}
p(\cdot) =
p(\boldsymbol{X}_{\mathcal{F}\setminus\mathcal{S}}).
\end{equation*}
This corresponds to $\texttt{do}$-intervention under a causal graph with no feature dependencies~\cite{janzing2020feature, heskes2020causal}. While it isolates the model's response to specific features, it frequently generates \textbf{off-manifold} inputs, i.e. combinations of values that could never occur in reality. This may lead to misleading attributions in highly correlated datasets~\cite{merrick2020explanation}.

\subsection{Joint-Marginal Shapley Values}

The \textit{joint-marginal} variant treats every absent feature as independent, sampling each from its univariate marginal~\cite{datta2016algorithmic}, i.e.
\begin{equation*}  
p(\cdot) =
\prod_{j\in\mathcal{F}\setminus\mathcal{S}} p(X_j).
\label{eq:jointmarginal}
\end{equation*}
By enforcing independence, it simplifies computation but amplifies the \textit{off-manifold} problem. In our study, this variant was rated as the \textbf{clearest} by analysts, suggesting that "true-to-the-model" logic may be easier for humans to parse than complex distributions.

\subsection{Conditional Shapley Values}

The \textit{conditional} formulation preserves empirical dependencies by conditioning on observed coalition values~\cite{aas2021explaining, chen2023algorithms}:
\begin{equation*}
p(\cdot) = p(\boldsymbol{X}_{\mathcal{F}\setminus\mathcal{S}} \mid \boldsymbol{X}_{\mathcal{S}}=\boldsymbol{x}_{\mathcal{S}}).
\label{eq:conditional}
\end{equation*}
This produces realistic \textit{in-manifold} imputations but is computationally intensive and difficult to estimate in high dimensions, where practical implementations rely on approximate matching, nearest neighbours, or parametric density models. Our audit shows this variant significantly \textbf{inflates analyst confidence} without improving accuracy.

\subsection{Uniform Shapley Values}

The \textit{uniform} formulation replaces absent features with samples drawn uniformly from a predefined hyperbox~\cite{strumbelj2010efficient, merrick2020explanation}:
\begin{equation*}
p(\cdot) = u(\boldsymbol{X}_{\mathcal{F} \setminus \mathcal{S}}).
\label{eq:uniform}
\end{equation*}
While useful for exploring the model's global behavior, it often results in noisy, confusing explanations for human reviewers.

\subsection{Search Counterfactual Shapley Values}

This formulation identifies the minimal changes needed to flip a model's prediction~\cite{albini2022counterfactual}. It replaces absent features with values from a vector $\boldsymbol{X}^{(c)}$ that reaches a target outcome $y^*$:
\begin{equation*}
\label{eq:counterfactual}
p(\cdot) = p(\boldsymbol{X}^{(c)}_{\mathcal{F}\setminus\mathcal{S}} \mid f(\boldsymbol{X}_{\mathcal{F}}^{(c)})\!\approx\!y^*),
\end{equation*}
This emphasizes \textbf{actionability} is but is computationally demanding~\cite{karimi2021algorithmic}, and can be sensitive to chosen distance metrics or plausibility constraints.  

\subsection{Filtered Conditional Shapley Values}

This variant restricts the background distribution to samples where model outputs fall within a specific range $\mathcal{Y}$:
\begin{equation*}
p(\cdot) = p(\boldsymbol{X}_{\mathcal{F}\setminus\mathcal{S}} \mid f(\boldsymbol{X}_{\mathcal{F}})\!\in\!\mathcal{Y}).
\label{eq:filteredconditional}
\end{equation*}
In our audit, this produced the most \textbf{sparse} and \textbf{contrastive} explanations, yet was consistently rated as confusing by professional analysts.

\subsection{Excluded Formulations} 

We excluded causal~\cite{heskes2020causal} and generative in-manifold~\cite{taufiq2023manifold} variants due to their high computational overhead and requirement for strong causal assumptions. These are currently incompatible with the millisecond-level latency required in production risk systems and are thus outside the scope of our operational audit.

\section{Properties of Explanation Quality}
\label{app:functional}

Recent taxonomies of explainability~\cite{canha2025functionally} identify a broad set of \emph{functionally grounded properties} that characterize a "good" explanation. These span computational, representational, and human-centered dimensions. This appendix refines these concepts for the \emph{Shapley value framework}, distinguishing between properties that are inherent to the framework and those that are formulation-dependent and thus central to our audit. Table~\ref{tab:criteria} provides a high-level overview.

\subsection{Properties Invariant Across Shapley Configurations}
\label{app:irrelevant}

The following properties are inherent to the Shapley framework itself, and remain constant across all eight variants in our study. Thus, they do not serve as discriminators in our evaluation.

\paragraph{Representativeness.} Shapley values are, by construction, \textit{local feature attributions} that are model-agnostic~\cite{lundberg2017unified, aas2021explaining}. All formulations share this scope; therefore, representativeness does not vary between definitions.

\paragraph{Structure.} The structural form of a Shapley explanation is always additive. Consequently, the visual format (e.g., bar charts) is independent of the value-function choice. All variants in our study were presented using the exact same visual encoding to ensure structural parity.

\paragraph{Interactivity.} Interaction depends on the UI (e.g., toggles, sliders) rather than the mathematical definition. By holding the interface fixed, we ensure interactivity remains a constant across our experimental matrix.

\paragraph{(Un)certainty.}
While uncertainty estimates can be attached to explanations (e.g., via ensemble bootstraps~\cite{covert2021explaining}), these quantify the stochasticity of the underlying model or the estimation procedure, not the conceptual choice of the value function.

\paragraph{Speed.} While training costs vary, the \emph{inference} speed is standardized by our use of amortized surrogates. This ensures that the operational feasibility of the explanations is consistent across all formulations during the analyst review process.

\subsection{Properties Relevant for Shapley Evaluation}
\label{app:relevant}

These properties depend directly on the choice of value function and form the basis of our empirical audit.

\paragraph{Selectivity.} This captures how concentrated attributions are across features, and determines the amount of information users must process, affecting cognitive load~\cite{canha2025functionally}. Different formulations produce different sparsity profiles. We quantify this via the \textit{Sparsity} metric~\eqref{eq:sparsity}, testing the assumption that "less is more" for human interpretability.

\paragraph{Contrastivity}
It captures whether explanations express differences relative to a meaningful reference or baseline~\cite{canha2025functionally}. Certain Shapley definitions (e.g., counterfactual, causal) are inherently contrastive. Other definitions may conflate average and specific contributions, leading to sensitivity to arbitrary baselines. We operationalize this through sensitivity to counterfactual perturbations~\eqref{eq:contrastivity}.

\paragraph{Fidelity} Fidelity reflects how accurately the additive decomposition of Shapley attributions reconstruct a black-box prediction. Specifically, how accurately an amortized surrogate reconstructs the "ground truth" for its specific definition. We use \textit{Attribution Error} and \textit{Recall@k} as control measures to ensure that our surrogates are mathematically sound.

\paragraph{Faithfulness} It captures whether the explanation preserves the causal influence of features on predictions~\cite{molnar2020interpretable}. Because different value functions define different intervention semantics, faithfulness varies significantly. We evaluate this using \textit{Deletion and Insertion AUC}~\eqref{eq:deletionauc}.

\paragraph{Truthfulness} Truthfulness connects the explanation to domain reality and expert priors~\cite{canha2025functionally}. Some value functions can yield explanations that contradict established domain knowledge. We assess this through our human-in-the-loop experiments in Section~\ref{sec:humanevaluation}, measuring whether an explanation helps or hinders an analyst's decision-making process.

\paragraph{Stability} Stability denotes reproducibility under small input or model perturbations. Formulations that integrate over correlated feature distributions tend to produce smoother attributions. We operationalize this via the \textit{Perturbation Sensitivity} metric~\eqref{eq:sensitivity}.

\section{Datasets}
\label{app:datasets}

We evaluate five datasets commonly used in risk assessment and explainability research, summarized in Table~\ref{tab:datasets}. These span healthcare, credit scoring, income prediction, and fraud detection, and vary in feature types, class imbalance, and task complexity.

\begin{table}[h]
    \caption{Summary statistics for experimental datasets; including sample sizes for training, validation, and A/B testing with amortisers. Prevalence refers to the proportion of positive (risk) instances.}
    \label{tab:datasets}
    \renewcommand{\arraystretch}{1.2}
    \setlength{\tabcolsep}{4.7pt}
    \begin{tabular}{lcccccc}
        \toprule
        \textbf{Dataset} & \textbf{Feat.} & \textbf{Cat.} & \textbf{Prev.} & \textbf{Train} & \textbf{Val.} & \textbf{Test} \\
        \midrule
        Maternal Risk & 6 & 0 & 26.8\% & 812 & 101 & 101 \\
        German Credit & 20 & 11 & 30.0\%  & 800 & 100 & 100 \\
        Adult & 10 & 6 & 63.1\% & 998{,}300 & 200 & 200 \\
        HELOC & 22 & 0 & 52.1\% & 7{,}029 & 200 & 200 \\
        Fraud Risk* & 54 & 15  & 9.1\% & 16{,}496  & 200 & 200 \\
        \midrule
        \multicolumn{7}{l}{\small *Heavily downsampled to achieve a 10-1 risk ratio before amortization.} \\
        \bottomrule
    \end{tabular}
\end{table}

\paragraph{Maternal Risk}
Clinical measurements used to predict maternal health outcomes~\cite{ahmed2020review}. The task is to predict whether a pregnancy is associated with elevated health risk. This represents a low-dimensional, fully continuous healthcare task.

\paragraph{German Credit}
The German Credit dataset~\cite{statlog_(german_credit_data)_144} consists of an anonymized financial and demographic set of features for loan applicants. It combines numerical and categorical attributes describing credit history and employment, frequently used to study the intersection of XAI and algorithmic fairness.

\paragraph{Adult}
The Adult (Income) dataset~\cite{ding2021retiring} contains demographic and employment-related attributes used to predict income thresholds. We use a curated subset of features from the Aequitas package~\cite{aequitas} to focus on socio-economic indicators common in automated decision-making.

\paragraph{HELOC}
Credit bureau attributes from FICO~\cite{fico2025educational-analytics} used to predict risk for home equity lines of credit. This dataset is a cornerstone for XAI evaluation in the financial sector due to its high-dimensional numerical feature set.

\paragraph{Fraud Risk}
This is a proprietary, large-scale dataset of card payment transactions collected and downsampled from a real-world financial risk assessment system. Each instance is represented by a heterogeneous set of numerical and categorical features capturing transactional, behavioral, and contextual information available at decision time. The dataset is highly imbalanced, reflecting real operational fraud prevalence.

\subsection{Dataset Preprocessing}
\label{app:preprocessing}

All datasets undergo a standardized preprocessing pipeline. Categorical features are ordinally encoded to preserve fixed dimensionality across models and explainers. Numerical features are used as provided. Dataset-specific preprocessing is minimal, except for HELOC, which contains missing or invalid sentinel values; these are handled via feature filtering, instance removal, and targeted imputation.

Each dataset is divided into training, validation, and test sets using stratified sampling to preserve class proportions. Validation and test sets are capped at 10\% of the dataset size or a maximum of 200 instances each for A/B Testing, whichever is smaller. All splits are generated using a fixed random seed to ensure reproducibility. The final number of features, label prevalence, and split sizes for each dataset are reported in Table~\ref{tab:datasets}.

\section{Extended Experimental Setup}
\label{app:model_training}

Our audit utilizes a standardized training and optimization pipeline to ensure that variations in explanation quality are not confounded by model performance or optimization noise.

We train two predictive binary classification risk models per dataset: Logistic Regression and LightGBM~\cite{ke2017lightgbm}. Hyperparameters were optimized via \textit{Optuna}~\cite{optuna} with a Tree-structured Parzen Estimator (TPE) sampler and median pruning. We use 3-fold stratified cross-validation and 20 trials per configuration. The optimization objective was set to validation AUC. Final model configurations are summarized in Table~\ref{tab:model_hyperparams}.

\begin{table}[ht!]
    \centering
    \caption{Risk model hyperparameter configurations.}
    \label{tab:model_hyperparams}
    \setlength{\tabcolsep}{2.2pt}
    \renewcommand{\arraystretch}{1.1}
    \begin{tabular}{lccccc}
        \toprule
        \textbf{Hyperparameter} & \textbf{Maternal} & \textbf{Credit} & \textbf{Adult} & \textbf{HELOC} & \textbf{Fraud} \\
        \midrule
        \multicolumn{6}{l}{\textit{LightGBM}} \\
        \cmidrule(lr){1-6}
        Num. Leaves & 11 & 82 & 98 & 59 & 6 \\
        Learning Rate & 0.27 & 0.10 & 0.14 & 0.05 & 0.10 \\
        N-Estimators & 258 & 121 & 285 & 187 & 50 \\
        Max. Depth & 4 & 4 & 6 & 3 & 3 \\
        Min. Child Samples & 25 & 22 & 90 & 100 & 500 \\
        Subsample & 0.76 & 0.65 & 0.60 & 0.61 & 0.50 \\
        Bagging Frac. & 0.59 & 0.68 & 0.96 & 0.99 & 0.50 \\
        Feature Frac. & 0.65 & 0.86 & 0.54 & 0.76 & 0.50 \\
        Colsample/tree & 0.72 & 0.61 & 0.52 & 0.89 & 0.50 \\
        Early Stopping & 72 & 132 & 78 & 179 & 20 \\
        \midrule
        \multicolumn{6}{l}{\textit{Logistic Regression}} \\
        \cmidrule(lr){1-6}
        C (Reg. Strength) & 0.13 & 0.36 & 0.60 & 0.59 & 0.09 \\
        Tol & 0.00 & 0.01 & 0.00 & 0.00 & 0.00 \\
        Solver & $S_1$ & $S_1$ & $S_2$ & $S_1$ & $S_3$ \\
        Max Iterations & 648 & 237 & 374 & 102 & 785 \\
        Class Weight & Bal. & Bal. & Bal. & Bal. & Bal. \\
        Fit Intercept & T & F & T & T & F \\
        Intercept Scaling & 0.66 & 0.82 & 0.31 & 1.53 & 0.19 \\
        \bottomrule
        \addlinespace[2pt]
        \multicolumn{6}{l}{\small $S_1$: newton-cholesky, $S_2$: sag, $S_3$: newton-cg, Bal.: Balanced}
    \end{tabular}
\end{table}

\paragraph{Amortizer Architecture and Training}
To support the low-latency requirements of the audit, we deploy feed-forward neural network amortizers.
\begin{itemize}
    \item \textbf{Preprocessing:} Numerical features are scaled using robust scaling (IQR-based) with \textit{tanh} saturation to mitigate outliers. Categorical features are passed through learned embedding layers.
    \item \textbf{Architecture:} Layers include fully connected blocks of varying dimensionality, Layer Normalization, LeakyReLU activations, and Dropout ($0.1$).
    \item \textbf{Optimization:} We employ a two-phase schedule: (1) Adam optimization with a linear warmup, followed by (2) SGD with momentum and cosine decay.
    \item \textbf{Parameters:} Amortizers are trained for $1000$ epochs ($100$ for Adult) with a learning rate of $0.001$. We sample four Shapley masks per input during training, with an efficiency loss weight of $0.1$.
\end{itemize}

\paragraph{Implementation of Value Functions}
We amortize eight distinct value functions (summarized in Appendix~\ref{app:shapleyformulations}). Sampling-based formulations (Marginal, Conditional, etc.) utilize a background set of $100$ instances specifically generated according to the respective sampling logic. Fixed baseline formulations are implemented using both Zero and Mean reference points to contrast their performance against empirical variants.

\section{Human-in-the-Loop A/B Testing Framework Details}
\label{app:abtesting}

To ensure the validity of our audit, we recruited a diverse pool of participants and professional analysts to interact with a standardized risk-assessment interface. By holding the UI, model outputs, and task context constant, we isolated the choice of Shapley formulation as the sole independent variable.

\subsection{Participant Profiles}

Table~\ref{tab:participants} summarizes the 37 individuals who contributed to the 3,735 human--AI interactions recorded in this study. The pool includes $5$ professional fraud analysts ($13.5\%$) and a high proportion of participants with significant ML and Shapley-specific knowledge, ensuring that the observed "confidence-inflation" was not merely a result of user naivety. Here, \textbf{domain knowledge} reflects self-assessed familiarity with the specific dataset being analyzed, presented separately for each dataset included in the study.

\begin{table}[h]
    \centering
    \caption{Summary of participant profiles in the A/B testing experiments.}
    \label{tab:participants}
    \setlength{\tabcolsep}{6pt}
    \renewcommand{\arraystretch}{1.1}
    \begin{tabular}{lccc}
        \toprule
        \textbf{Total participants} & \multicolumn{3}{c}{37} \\
        \textbf{Professional analysts} & \multicolumn{3}{c}{5 (13.5\%)} \\
        \midrule
         & \textbf{Low} & \textbf{Moderate} & \textbf{High} \\ 
        \textbf{ML knowledge} & 9 (24.3\%) & 10 (27.0\%) & 18 (48.6\%) \\
        \midrule
        \multicolumn{2}{c}{} & \textbf{Yes} & \textbf{No} \\
        \multicolumn{2}{l}{\textbf{Shapley knowledge}} & 25 (67.6\%) & 12 (32.4\%) \\
        \midrule
        \multicolumn{2}{l}{\textbf{Domain knowledge}} & \textbf{Yes} & \textbf{No} \\
        \multicolumn{2}{l}{Maternal Risk} & 3 (12.5\%) & 21 (87.5\%)  \\
        \multicolumn{2}{l}{UCI German Credit}  & 4 (13.3\%) & 26 (86.7\%) \\
        \multicolumn{2}{l}{Adult} & 3 (12.5\%) & 21 (87.5\%) \\
        \multicolumn{2}{l}{FICO HELOC}  & 2 (10.0\%) & 18 (90.0\%) \\
        \multicolumn{2}{l}{Fraud Risk}  & 8 (38.1\%) & 13 (61.9\%) \\
        \bottomrule
    \end{tabular}
\end{table}

\subsection{Experimental Interface and Task Flow}
\label{app:ui}

The interface was designed to resemble operational risk analysis tools commonly used in practice. All visual components, layouts, and interaction patterns were held fixed across datasets, models, and Shapley value formulations.

\subsubsection{Overview and task flow}

The experiment followed a strictly linear workflow to prevent backtracking or cross-case comparison:
\begin{enumerate}
    \item \textbf{Onboarding:} Dataset selection and completion of the analyst profile.
    \item \textbf{Contextualization:} Review of standardized, dataset-specific task summaries.
    \item \textbf{Review Loop:} Sequential case analysis and decision-making.
\end{enumerate}
Participants could only advance after submitting a final decision, ensuring that recorded interactions were independent.

\subsubsection{Dataset Selection and Analyst Profile}

\begin{figure*}[t!]
    \centering
    \includegraphics[width=0.98\linewidth]{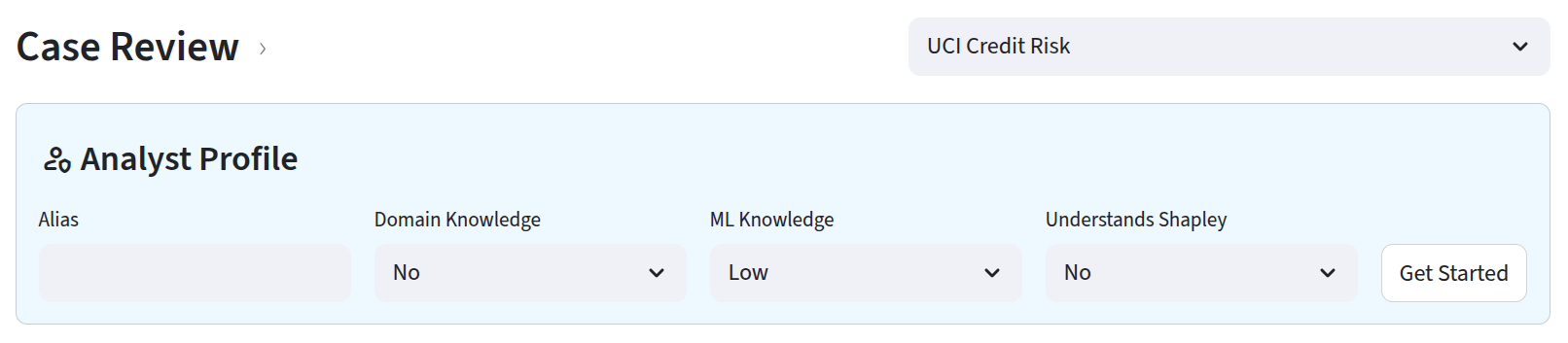}
    \caption{A screenshot showing the initial onboarding form with dropdowns for expertise levels.}
    \label{fig:profile}
\end{figure*}

Upon selecting a dataset from a drop-down menu, participants used a panel to self-report their expertise in three areas: domain knowledge for the specific data, general ML familiarity, and prior experience with Shapley-based explanations.  These attributes are collected exclusively for post-hoc analysis and stratification and do not influence task assignment, model selection, or explanation configuration. A screenshot of the dataset selection and analyst profile form is shown in Figure~\ref{fig:profile}.

\subsubsection{Task summary panel}

\begin{figure}[H]
    \centering
    \includegraphics[width=1\linewidth]{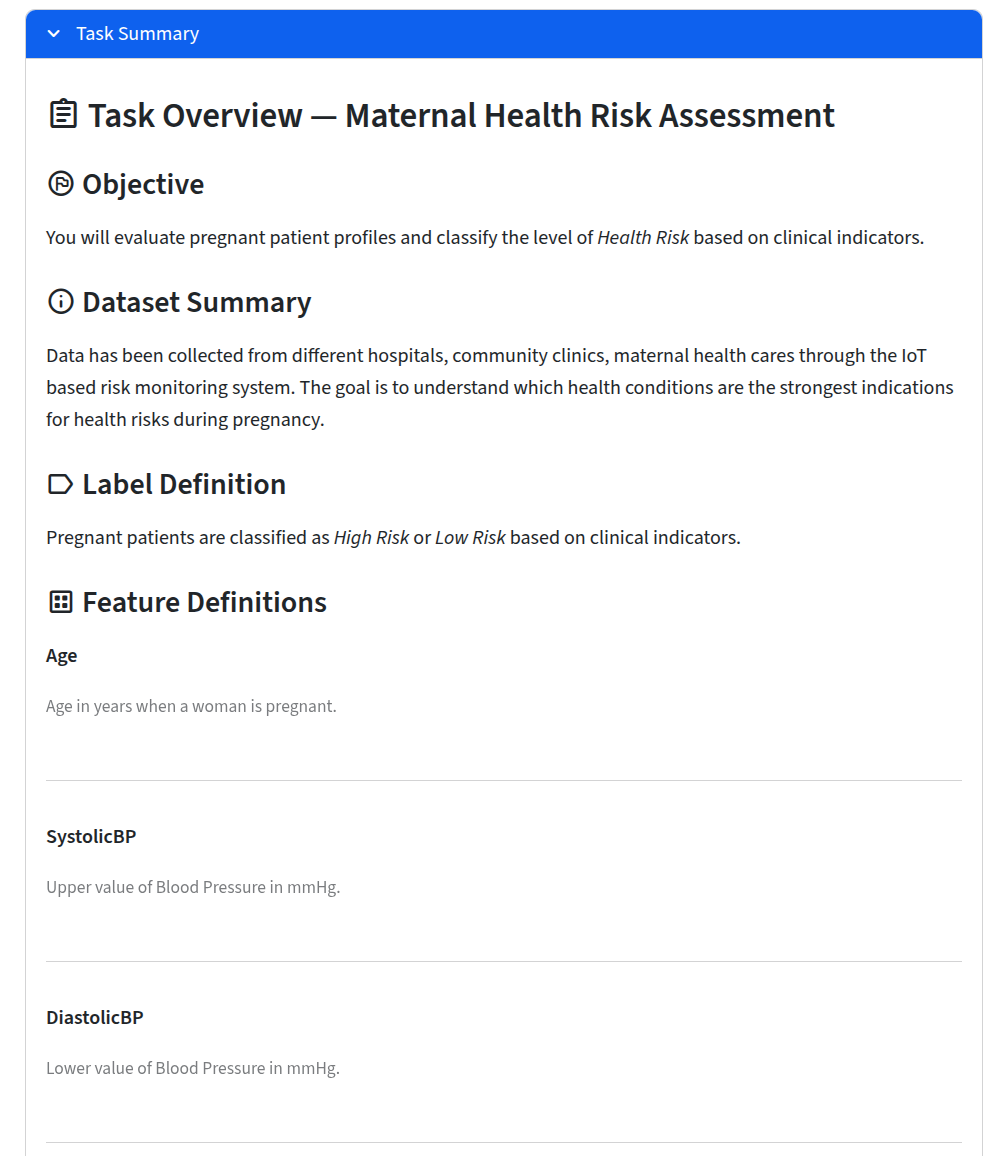}
    \caption{Task summary panel providing analysts with dataset and task context.}
    \label{fig:tasksummary}
    \Description{}
\end{figure}

A collapsible summary panel provided a constant reference for the prediction objective, feature definitions (including categorical levels), and the operational meaning of \textit{risk}. This ensured that all participants, regardless of prior expertise, operated under a shared conceptual framework.  Figure~\ref{fig:tasksummary} shows an example of the task summary panel.

\subsubsection{Case-review screen}

The interface (detailed in Figure~\ref{fig:main_casereview} of the main text) comprised three primary information blocks:

\paragraph{Model output.}
The risk score is displayed as a value in $[0,1]$, together with its empirical percentile within the dataset. A score distribution plot provides additional context by situating the instance relative to the overall population.

\paragraph{Explanation panel.}
Explanations featured a horizontal bar chart of the four features with the highest absolute Shapley attributions, using consistent color, scale, and ordering conventions across all Shapley formulations. To mirror real-world compliance requirements, the interface displays automatically generated \emph{reason codes}, i.e. short natural-language statements deterministically derived from the Shapley values (e.g., ``\textit{Age value of 70 is high}'').

\paragraph{Data explorer.}
A read-only explorer allows participants to inspect the raw feature values of the current instance. Numerical features were accompanied by univariate distributions, while categorical features included prevalence and historical risk statistics. This ensured analysts could verify the data without manipulating it.

\subsubsection{Decision and feedback collection}
\begin{figure}[H]
    \centering
    \begin{subfigure}[t]{\linewidth}
        \centering
        \includegraphics[width=0.9\linewidth]{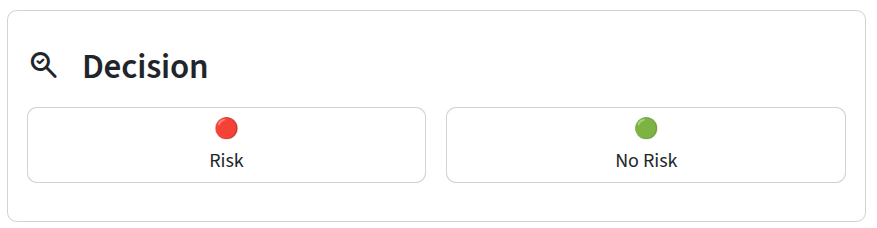}
        \label{fig:decision}
    \end{subfigure}
    \hfill
    \begin{subfigure}[t]{\linewidth}
        \centering
        \includegraphics[width=0.9\linewidth]{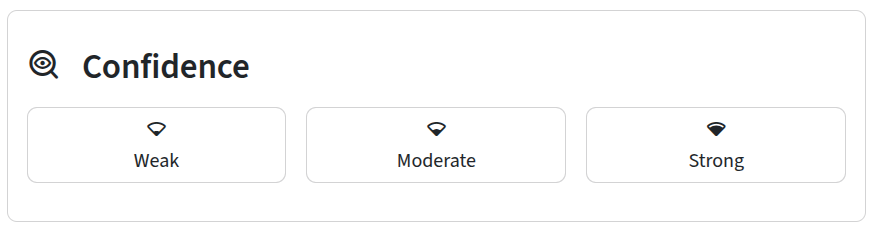}
        \label{fig:confidence}
    \end{subfigure}
    \begin{subfigure}[t]{\linewidth}
        \centering
        \includegraphics[width=0.9\linewidth]{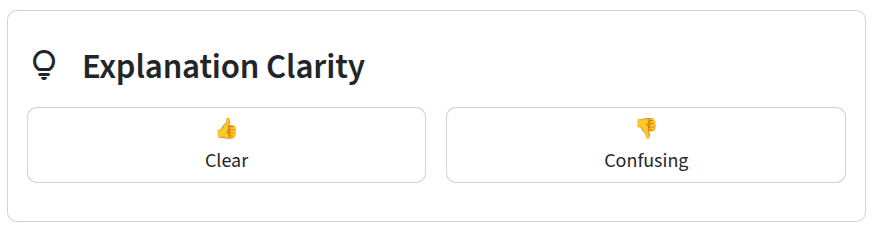}
        \label{fig:clarity}
    \end{subfigure}
    \hfill
    \caption{Sequence of questions asked to the analyst for each reviewed case.}
    \label{fig:decision_sequence}
\Description{}
\end{figure}

The feedback loop (Figure~\ref{fig:decision_sequence}) was designed to capture the core outcome variables of our audit. After reviewing a case, analysts submitted:
\begin{itemize}
    \item \textbf{Decision:} A binary "Risk" or "No Risk" judgment.
    \item \textbf{Confidence:} Self-reported certainty (\textit{Weak, Moderate, Strong}).
    \item \textbf{Clarity:} A subjective assessment of the explanation (\textit{Clear} or \textit{Confusing}).
\end{itemize}

No additional controls, filters, or explanation parameters are exposed. This design ensures that observed differences in decision behavior, confidence, and response time arise from explanation content rather than interface affordances or user-driven customization.

\end{document}